\documentclass[10pt,twocolumn,letterpaper]{article}

\usepackage[pagenumbers]{cvpr} 

\usepackage[dvipsnames]{xcolor}

\definecolor{cvprblue}{rgb}{0.21,0.49,0.74}
\usepackage[pagebackref,breaklinks,colorlinks,citecolor=cvprblue]{hyperref}

\def\confName{CVPR}
\def\confYear{2024}
\def\OURS{\textsc{ger}\xspace}
\def\OURSf{\textsc{ger-ald}\xspace}
\def\OURSa{\textsc{ger-atomic}\xspace}
\def\OURSc{\textsc{ger-caption}\xspace}
\def\OURSh{\textsc{ger-hkc}\xspace}

\definecolor{aliceblue}{rgb}{0.94, 0.97, 1.0}
\definecolor{mistyrose}{rgb}{1.0, 0.89, 0.88}
\definecolor{Gray}{gray}{0.9}
\definecolor{emerald}{RGB}{80,200,120}
\definecolor{cerulean}{RGB}{42,82,190}
\definecolor{red_plot}{RGB}{199,100,38}
\definecolor{blue_plot}{RGB}{47, 114, 173}
\newcommand{\green}[1]{\textcolor{emerald}{#1}}
\usepackage{bbding}
\usepackage{pifont}
\usepackage{stix}
\usepackage{soul}
\usepackage[ruled,linesnumbered,algo2e]{algorithm2e}
\usepackage{amsfonts}
\usepackage{stmaryrd}
\usepackage{paralist}
\usepackage{algorithm}
\usepackage{algpseudocode}
\usepackage{xcolor,colortbl}

\newcommand\blfootnote[1]{%
  \renewcommand\thefootnote{}\footnote{#1}%
  \addtocounter{footnote}{-1}%
}

\title{A Generative Approach for Wikipedia-Scale Visual Entity Recognition}

\author{Mathilde Caron
~~~~~Ahmet Iscen
~~~~~Alireza Fathi
~~~~~Cordelia Schmid
\\
Google Research
}

\begin{document}
\maketitle
\begin{abstract}

\blfootnote{
\hspace{-0.55cm}
Code: \href{https://github.com/google-research/scenic/tree/main/scenic/projects/gerald}{github.com/google-research/scenic/tree/main/scenic/projects/gerald}
}
In this paper, we address web-scale visual entity recognition, specifically the task of mapping a given query image to one of the 6 million existing entities in Wikipedia. 
One way of approaching a problem of such scale is using dual-encoder models (\eg CLIP), where all the entity names and query images are embedded into a unified space, paving the way for an approximate $k$NN search.
Alternatively, it is also possible to re-purpose a captioning model to directly generate the entity names for a given image. 
In contrast, we introduce a novel \underline{G}enerative \underline{E}ntity \underline{R}ecognition
(\OURS) framework, which given an input image learns to auto-regressively decode a semantic and discriminative ``code'' identifying the target entity.
Our experiments demonstrate the efficacy of this \OURS paradigm, showcasing state-of-the-art performance on the challenging OVEN benchmark. 
\OURS surpasses strong captioning, dual-encoder, visual matching and hierarchical classification baselines, affirming its advantage in tackling the complexities of web-scale recognition.

\end{abstract}    
\section{Introduction}
\label{sec:intro}

Generative vision-language models such as GPT-4~\cite{gpt4}, Flamingo~\cite{alayrac2022flamingo} or PALI~\cite{pali2022}, are becoming increasingly popular for computer vision applications.
They show an impressive ability to generate free-form text for describing the contents of an image (captioning), or answering questions based on an image (visual-question answering). 
Nevertheless, their potential for \emph{recognition} tasks~\cite{hu2023open}, which usually require a more concise, structured output, remains under-explored. 
The focus of this paper is to explore their application for the challenging task of web-scale entity recognition.
A recent benchmark, Open-domain Visual Entity recognitioN (OVEN)~\cite{hu2023open}, challenges models to associate an image with a Wikipedia entity from a pool of over six million entities.
Models must establish a robust association between images across millions of coarse-grained and fine-grained entities, encompassing a wide spectrum of concepts such as animals, buildings, locations, and a multitude of others~\cite{hu2023open}.

\begin{figure}[t]
\centering
\includegraphics[width=0.95\linewidth]{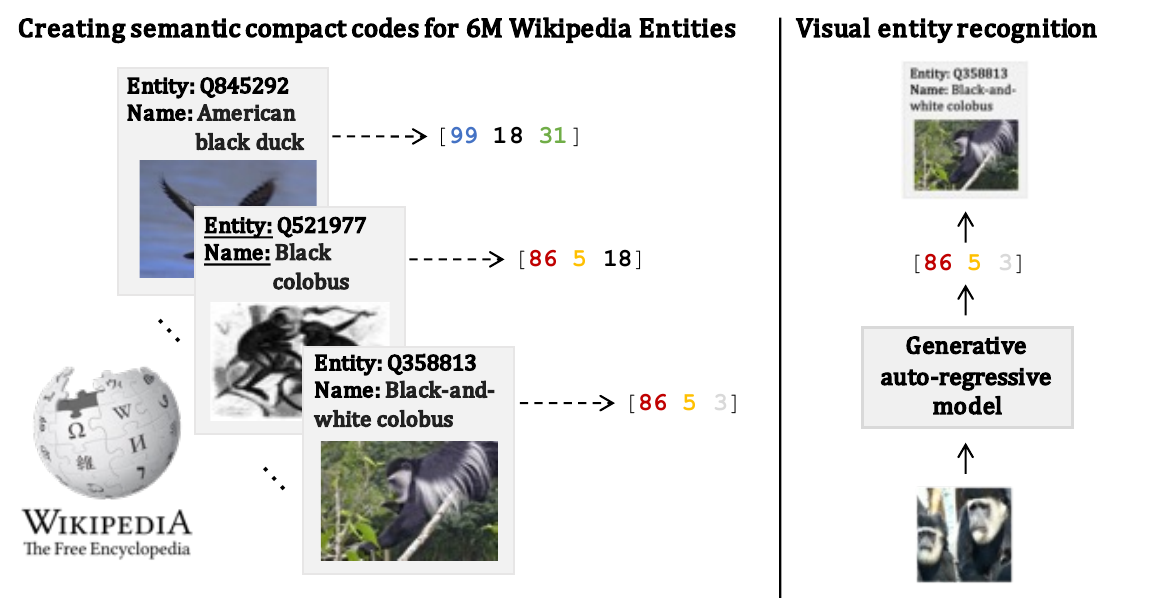}
\vspace{-0.2cm}
\caption{We introduce \OURS, a novel generative paradigm for web-scale visual entity recognition. 
We create compact semantic codes for each entity, and learn to auto-regressively generate them for a given query image at inference.
\vspace{-0.4cm}
}
\label{fig:teaser}
\end{figure}

Traditionally, the predominant methods employed to address the challenge of visual entity recognition have revolved around either classification or contrastive dual-encoder paradigm like CLIP~\cite{radford2021learning}.
While classification offers a straightforward approach, it grapples with limitations when confronted with extensive label spaces such as that of OVEN, resulting in substantial parameter counts and practical engineering complexities.
The dual-encoder approach on the other hand, learns a unified image-text feature space, thereby facilitating efficient nearest neighbor searches for recognition.
Nonetheless, this approach exhibits its own drawbacks: (a) it does not directly optimize for the final recognition task but instead relies on indirect optimization through contrastive loss where a set of negative data has to be subsampled at training time~\cite{gutmann2010noise,oord2018representation,radford2021learning}, (b) compressing either the image or text into an embedding vector results in loss of information, detrimentally affecting performance for fine-grained recognition~\cite{iscen2023retrieval} and (c) the memory requirements for storing dense representations scale proportionally with the size of the entity set.

These challenges of the dual-encoder paradigm have kindled interest in alternative strategies.
Notably, in Natural Language Processing (NLP) domain, recent works challenge the dual-encoder approach and use generative models instead for information retrieval~\cite{tay2022transformer,rajput2023recommender,pradeep2023does,de2020autoregressive,mehta2022dsi++,sun2023learning}.
These works represent each element of the corpus by a compact code of integers, and learn an auto-regressive generative model to decode the target code for a given query.
This paradigm promises to overcome some drawbacks of dual-encoders by simplifying the retrieval pipeline such that the training and inference objectives are the same, and directly encoding the corpus within the model's parameters.
Also as an alternative to dual encoders, OVEN paper~\cite{hu2023open} showcases the feasibility of extending a generative image captioning model~\cite{pali2022} for visual entity recognition by matching the generated caption to one of the Wikipedia entity texts~\cite{robertson2009probabilistic}.

Inspired by these recent explorations, we propose a \underline{G}enerative \underline{E}ntity \underline{R}ecognition (\OURS) framework (illustrated in Fig.~\ref{fig:teaser}) to facilitate end-to-end visual entity recognition by leveraging generative auto-regressive models.
Specifically, we represent each Wikipedia entity with a \textit{code}, \ie a short sequence of integers.
Then, we train models to predict an entity from an input image by auto-regressively generating the code corresponding to the target entity.
We find that creating un\underline{A}mbiguous, \underline{L}anguage-based and \underline{D}iscriminative (\textsc{ald}) entity codes results in the best variant of our \OURS framework, which we denote by \OURSf.
In fact, while we observe that unstructured ``atomic'' codes work well in some scenarios, they fail when training data or model capacity are limited or more importantly, when the entity set reaches the million scale (see Sec.~\ref{sec:atomic}).
Plus, they cannot generalize to new entities.
In contrast, we find that semantically-structured codes based on language improve upon atomic codes by leveraging generic concepts shared across related entities (see example in Fig.~\ref{fig:teaser} with ``Black colobus'' and ``Black-and-white colobus'' sharing common code tokens).
A simple way of creating codes based on language is to directly tokenize~\cite{kudo2018sentencepiece} the entity name, which is akin to image captioning where the entity name is used as a caption~\cite{hu2023open,de2020autoregressive}.
However, we find that such tokenized entity names contain clutter and noisy information, all the more so when the entity name is long (see Sec.~\ref{sec:captioning}).
Our \OURSf method improves over this simple captioning baseline by decoding only the most \textit{discriminative} part of the tokenized entity name, \ie the part which makes the considered entity name the most different compared to all other entities.

Finally, we also propose an entity-based pre-training to condition the \OURS models to web-scale entity recognition.
Inspired by recent advances in retrieval-based methods~\cite{iscen2023retrieval,liu2023learning}, we retrieve a subset of images from a large-scale image-text dataset typically used for captioning or contrastive pre-training~\cite{pali2022} and re-purpose it by replacing the original text captions with related OVEN entity names.
Overall, our experiments demonstrate the efficacy of the proposed \OURS paradigm: \OURSf outperforms previously published numbers on OVEN benchmark~\cite{hu2023open} by $+6.7$ top-1 accuracy, while using $42 \times$ less parameters.
In summary, our contributions are as follows:
\begin{compactitem}
\item a \underline{g}enerative \underline{e}ntity \underline{r}ecognition framework (\OURS) to facilitate end-to-end visual entity recognition;
\item  an innovative strategy for encoding Wikipedia entities into un\underline{a}mbiguous \underline{l}anguage-based \underline{d}iscriminative (\textsc{ald}) codes that are highly effective for \OURS;
\item an entity-based pre-training process without requiring human intervention;
\item state-of-the-art results in challenging web-scale OVEN entity recognition and on-par performance to traditional classifiers in smaller-scale label-space scenarios.
\end{compactitem}

\section{Related work}

\noindent\textbf{Visual entity recognition} aims to recognize classes, or entities given visual inputs~\cite{russakovsky2015imagenet}.
Granularity of visual entity recognition tasks varies from every-day generic objects~\cite{everingham2010pascal,fei2004learning}, to fine-grained domains, such as birds~\cite{wah2011caltech},  dogs~\cite{KhoslaYaoJayadevaprakashFeiFei_FGVC2011}, cars~\cite{krause20133d}, 
 food~\cite{bossard2014food}, landmarks~\cite{weyand2020google}, faces~\cite{zhu2022webface260m} and natural world species~\cite{van2018inaturalist}.
Some challenges for the visual entity recognition tasks include imbalanced training classes following a long-tailed distribution~\cite{OLTR}, or noisy training labels~\cite{li2017webvision}.
Recent work~\cite{hu2023open} proposes a new, \textit{web-scale} dataset for open-domain entity recognition.
This challenging benchmark contains $6$M entity names derived from Wikipedia page titles, including coarse-grained and fine-grained entities, encompassing a wide spectrum of concepts such as animals, buildings, organizations, landmarks, and a multitude of other.
The authors show that generative captioning models (\ie PaLI~\cite{pali2022}) outperform dual encoder models for large-scale entity recognition.
In this paper, we build upon this observation, and study generative models for accurate and efficient entity recognition.

\noindent\textbf{Extreme classification} tackles entity recognition specifically at a very large scale with a pure classification approach~\cite{bengio_et_al:DagRep.8.7.62,mittal2022multi,agrawal2013multi}.
Typical approaches explore strategies for scaling to the hundred of thousands scale and preliminary results are even shown at million scale~\cite{agrawal2013multi}.
By leveraging generative image-to-text models, we propose a fresh perspective beyond traditional classification methods typically used in the context of large-scale visual entity recognition.

\noindent\textbf{Generative auto-regressive retrieval} methods are increasingly popular in NLP~\cite{tay2022transformer,rajput2023recommender,pradeep2023does,de2020autoregressive,mehta2022dsi++,sun2023learning}.
GENRE retrieves Wikipedia entities by generating their names in an autoregressive fashion.
Seminal work DSI~\cite{tay2022transformer} shows the benefit of learning to decode compact codes (created either randomly or with hierarchical k-means clustering) associated with each document. 
Neural Corpus Indexer~\cite{wang2022neural} proposes a specific decoding scheme for generative retrieval and show the benefit of \emph{query augmentation} by automatically generating training queries for documents to be indexed.
TIGER~\cite{rajput2023recommender} studies generative retrieval in the context of recommender systems.
Finally, \cite{pradeep2023does} conducts a systematic study of generative retrieval systems when scaled to millions of document passages.
Only very few works explore this family of approaches in computer vision domain, and only in very small-scale and uni-modal scenarios~\cite{zhang2023irgen}.
 \section{Method}
\label{sec:method}
\begin{figure}[t]
\centering
\includegraphics[width=1\linewidth]{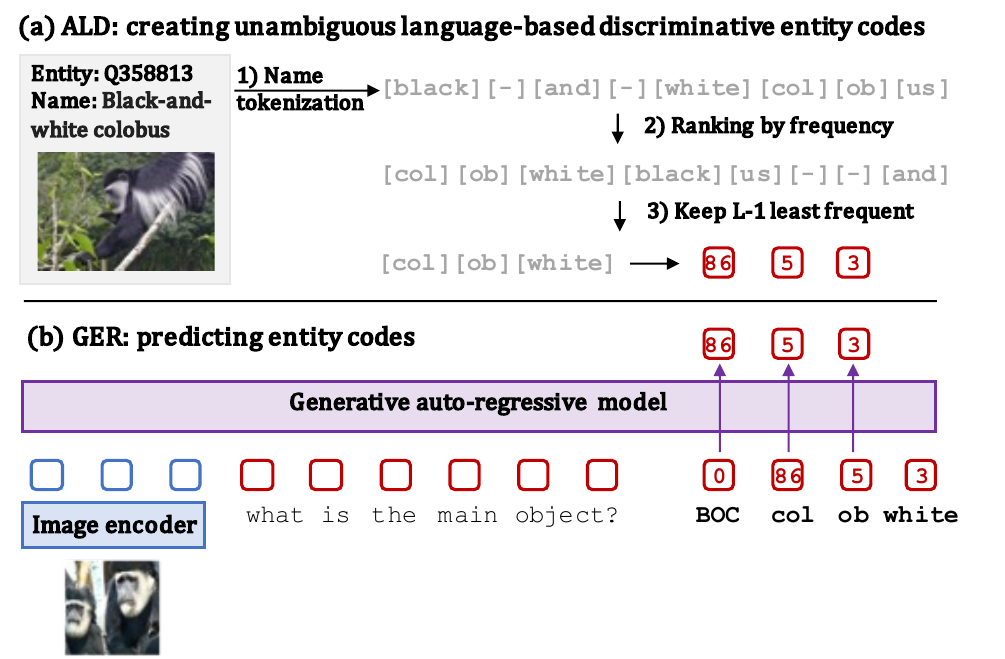}
\vspace{-0.6cm}
\caption{\textbf{Overview of \OURSf method.} \textbf{(a)} We utilize a text tokenizer to create compact and semantic codes, which represents each entity with short, but discriminative representations. \textbf{(b)} We learn a generative auto-regressive model, which learns to decode the correct code for given query image and text pair.
}
\vspace{-0.4cm}
\label{fig:method}
\end{figure}

Our goal is to explore how to adapt \underline{g}enerative auto-regressive models to the task of visual \underline{e}ntity \underline{r}ecognition (\OURS).
While previous works have shown preliminary signal that it is possible to repurpose autoregressive models for entity recognition by directly decoding entity names~\cite{hu2023open,de2020autoregressive}, we propose a more effective strategy.
An overview of our framework is in Fig.~\ref{fig:method}.

\subsection{Problem definition}
\label{sec:problem}

\noindent\textbf{Web-scale visual entity recognition.}
The Open-domain Visual Entity recognitioN (OVEN)~\cite{hu2023open} task consists of mapping input visual queries to one of the $6$M English Wikipedia entities.
More specifically, for a given image query $x_v$ and text query $x_t$, the model needs to recognize the corresponding entity $e$ among the set $\mathcal{E}$ of all possible entities.
The purpose of the input text $x_t$ is to achieve unambiguous recognition.
For example, when several entities are represented in the query image $x_v$, the text query indicates which one needs to be recognized.
Each entity $e \in \mathcal{E}$ comes with an entity name, denoted by $t_e$, which corresponds to the title of the entity Wikipedia page.

\noindent\textbf{Representing each entity with a code.}
 In \OURS, we represent each entity $e$ by a \textit{code} denoted by $c^e = \{c^e_1, ..., c^e_L\} \in \llbracket 1, V \rrbracket^L$ where $L$ is the length of the code and $V$ is the size of the \textit{vocabulary} of all integer values that each code token $c^e_i$ can take.
 This forms up to $V^L$ unique codes.
Note that vanilla image classification and captioning baselines can both be cast into this code formulation.
In fact, with $L=1$ and $V = |\mathcal{E}|$, the codes are equivalent to the labels used in standard multi-class classification.
On the other hand, if each code token value in $\llbracket 1, V \rrbracket $ maps to a (sub-)word in a pre-defined vocabulary~\cite{kudo2018sentencepiece}, then the codes simply correspond to standard tokenized text used in captioning models~\cite{kudo2018subword, sennrich2015neural,wang2022git}.
In the following paragraphs, we detail \OURSf, our most effective strategy for building codes $\mathcal{C}$ to represent all 6M English Wikipedia entities.

\subsection{\OURSf: Creating \textsc{ald} codes for \OURS}
\label{sec:codes}
We design the code set $\mathcal{C}$ so that it has three properties which we find are important for effective \OURS models:
i) semantically structured thanks to language, ii) discriminative and compact, and iii) unambiguous.
Our algorithm to create such un\underline{a}mbiguous, \underline{l}anguage-based and \underline{d}iscriminative codes, called \textsc{ald}, is illustrated in Fig.~\ref{fig:method} (a) and described in pseudo-code in Algorithm~\ref{alg} of the Appendix.

\noindent\textbf{Semantic tokens based on language.}
We find that entity codes $\mathcal{C}$ benefit from following a semantic structure, especially in scenarios where memorizing unstructured atomic codes is difficult.
We show in Sec.~\ref{sec:atomic} that using unstructured atomic codes fail when the amount of training data or the model capacity are limited or, of particular interest, when the entity set size increases to the million scale (see Fig.~\ref{fig:size_label_space}).
Intuitively, we want entities that are semantically similar to have some overlapping code tokens.
For example, we wish that entities $e = \small{\text{Q521977}}$ with corresponding name $t_{\text{Q521977}} = $ ``Black colobus'' and $e = \small{\text{Q358813}}$ with corresponding name $t_{\text{Q358813}} = $ ``Black-and-white colobos'' to share some code tokens, given that these correspond to two close species.

A simple yet effective way of having semantic codes is to tokenize the entity names based on text tokenizers~\cite{kudo2018subword, sennrich2015neural, kudo2018sentencepiece,de2020autoregressive}.
If each of the sub-words in the entity names are mapped to an integer representing this sub-word, then entities $\small{\text{Q358813}}$ and $\small{\text{Q521977}}$ naturally share code tokens: those representing the phrase ``colobus''.
We denote by $\Phi(.)$ an off-the-shelf text tokenizer with a vocabulary of $V_{\Phi}$ sub-words such that $\Phi(t_e) = \{y^e_1, ..., y^e_{L_e}\} \in \llbracket 1, V_{\Phi} \rrbracket^{L_e}$ where $L_e$ is the length of the tokenized entity name $\Phi(t_e)$.
In practice we use the same language tokenizer as GIT~\cite{wang2022git} for $\Phi(.)$ and have a vocabulary size of $V = V_{\Phi} = 30522$.
We refer to the baseline of using codes $\mathcal{C}$ created by simple tokenization of the entity name as \OURSc (i.e. we treat the entity name as a caption)~\cite{de2020autoregressive}.
We show in the following paragraph how \OURSf codes differ from such \OURSc codes by making them more compact and discriminative. 

\noindent\textbf{Discriminative and compact codes.}
Our goal is to build short and highly discriminative codes because they are easier to learn for the model, as validated by our experiments in Sec.~\ref{sec:captioning}.
For example, the tokenized entity name $\Phi(t_{\text{Q358813}}) = $ $\Phi(\text{``Black-and-white colobus''})$ counts $L_{\text{Q358813}} = 8$ tokens, but clearly not all $8$ tokens are important to make this entity discriminative compared to all other existing entities.
Hence, we choose to represent each entity with the \emph{bare minimum}, removing all the \emph{clutter} which is not only non-discriminative but also adds noise.
We achieve this by selecting the most discriminative and rarest tokens within the tokenized entity name.
Specifically, we compute the frequency $f_v$ of each token value $v \in [1, V]$ in the vocabulary over the entire corpus of tokenized entity names $\{\Phi(t_e)\}_{e \in \mathcal{E}}$.
We have $f_v = \frac{n_v}{\sum_{u=1}^V n_u}$ where $n_v$ is the number of times $v$ appears in $\{\Phi(t_e)\}_{e \in \mathcal{E}}$.
We create an \textsc{ald} code $c_e$ for each entity by keeping only the $(L-1)$ tokens with the lowest frequencies and discarding the other ones.
For example for entity $\small{\text{Q358813}}$, the 3 tokens with the lowest frequencies are``col'', ``ob'' and ``white''.
Interestingly, these 3 most discriminative tokens appear at the end of the code for \OURSc.
By contrast, they appear right at the beginning of the code for \OURSf and they constitute the only tokens to be decoded by the model, which intuitively explains the improved performance of \OURSf codes, as analyzed later in Sec.~\ref{sec:captioning} especially when entities have long names (see Fig.~\ref{fig:vis_examples}).
Finally an interesting by-product of using short codes is that they are faster to decode (the complexity of decoding is $\mathcal{O}(L^2)$) and require less memory footprint to store.

\noindent\textbf{Unambiguous codes.}
Note that several entities might share the same least frequent $(L-1)^{\text{th}}$ tokens.
In this case their code are exactly identical up to the $(L-1)^{\text{th}}$ token.
We use the last $L^{\text{th}}$ token to ensure that each entity has a unique code: we greedily assign the last code token $c^e_L$ to the next least frequent word of the tokenized entity name until the code $c_e$ is different from all existing codes.
If this still fails to create a unique code, we assign $c^e_L$ to a random token value $v'$ so that the resulting code is unique.
With code length $L=4$, only $0.5\%$ of the entities use a random token value.

 \subsection{Training}
\label{sec:training}
In this section, we describe the model used to decode entity codes from an input image-text pair.
Importantly, we also introduce our entity-based pre-training to condition the generative model to the task of entity recognition.

\noindent\textbf{Auto-regressive generative models.}
We build upon GIT~\cite{wang2022git}, an auto-regressive image-to-text generative model.
The query image-text pair $(x_v, x_t)$ is transformed into a set of $d$-dimensional embeddings using a visual encoder for $x_v$ and the text tokenizer $\Phi(.)$ for $x_t$.
The resulting output is represented by $\mathbf{X}_v \in \mathbb{R}^{N_v \times d}$ (resp. $\mathbf{X}_t \in \mathbb{R}^{N_t \times d}$) for image (resp. text) tokens.
We then input $\mathbf{X}_v$ and $\mathbf{X}_t$ to a decoder network $g(.)$ whose task is to decode the next code token $c^e_i$, conditioned on the previous tokens $c^e_{j<i}$.
Each code token value $v$ in $\llbracket 1, V \rrbracket$ maps to a learnable $d$-dimensional vector $\mathbf{Y}_v$ (gathered in the embedding matrix $\mathbf{Y} \in \mathbb{R}^{(V+1) \times d}$ where $\mathbf{Y}_0$ corresponds to the ``beginning of code'' token).
We train with a language modeling loss:
\vspace{-0.3cm}
$$
\mathcal{L}^e = \frac{1}{L} \sum_{i=1}^{L} \ell(c^e_i, g([\mathbf{X}_v; \mathbf{X}_t; \mathbf{Y}_{0}; \mathbf{Y}_{c^e_{0 < j < i}}]))
\vspace{-0.3cm}
$$
where $[;]$ corresponds to the concatenation operation in the first dimension and $\ell$ is the softmax cross-entropy loss with label-smoothing~\cite{muller2019does}.
We average $\mathcal{L}^e$ over a mini-batch and learn the weights of the visual encoder, decoder $g(.)$ and embedding matrix $\mathbf{Y}$ through back-propagation.
When decoding, we use beam search to obtain the best predicted entity coded.
We find that we do not need to constrain the beam search to existing codes since more than $99\%$ of the top-$1$ predictions are valid codes for converged \OURS models.

\noindent\textbf{Entity-based pre-training.}
Common auto-regressive models such as GIT~\cite{wang2022git} or PaLI~\cite{pali2022} are pre-trained for descriptive captioning.
As shown in~Tab.~\ref{ap:tab:captioning_zeroshot} and Fig~\ref{ap:fig:zeroshot_finetuned} of the Appendix, they generalize poorly to entity recognition.
This is because of the task discrepancy between predicting a descriptive caption and predicting an entity name.
In order to condition our models better for entity recognition, we propose to collect a significant number of entity-based pretraining images, each associated with a Wikipedia entity instead of a generic caption.
However, such an entity-based pretraining dataset does not exist.
We create it in an automatic way, without any human supervision.

To do so, we leverage existing large-scale image-caption datasets~\cite{schuhmann2021laion,schuhmann2022laion}: unless specified otherwise we use WebLI~\cite{pali2022}.
For each Wikipedia entity, we retrieve in WebLI the image-caption pairs that best represent this entity and replace their original captions by this entity name~\cite{iscen2023retrieval,liu2023learning}.
Specifically, we embed the $6$M entity names of OVEN with a semantic text encoder~\cite{radford2021learning} and find the top-$k$ most similar captions in WebLI.
We retrieve their corresponding images and replace their original captions by the considered entity name.
We ensure that no image is assigned to multiple entities to avoid instability during training.
We vary the number of retrieved images $k$ per entity from $2$ to $100$ to produce pre-training datasets of different sizes: from 11M up to 55M images (see Fig.~\ref{fig:pretraining}).
We denote by Entity-WebLI (resp. Entity-LAION) the resulting dataset used for entity-based pretraining, built from WebLI (resp. LAION~\cite{schuhmann2022laion}).
This way of creating pre-training data is akin to the query generation techniques used for generative retrieval in NLP~\cite{wang2022neural}.
However, rather than generating a synthetic input, we simply retrieve input images from a large-scale dataset.

 \subsection{Baselines}
\label{sec:baselines}
We compare our method to the following different baselines.

\noindent\textbf{Hierarchical classification.}
Solving million-scale entity recognition with classification is unpractical due to the very large number of classes.
A workaround is to use \textit{hierarchical} classifiers.
As OVEN does not come with hierarchical labels we obtain a 3-level hierarchy through k-means of the 6M entity names encoded with sentence-T5~\cite{ni2021sentence}.
We train a multi-class classifier for each parent node in the hierarchy.
To avoid training a huge number of different classification matrices, we learn a generic classifier matrix per level which is modified by learnable small \textit{modifiers} depending on the path in the hierarchy.

\noindent\textbf{Dual encoders.}
Another typical workaround to classification is to rely on deep metric learning approaches~\cite{schroff2015facenet} such as Noise Contrastive Estimation~\cite{gutmann2010noise} and its InfoNCE variant~\cite{oord2018representation} as used in popular dual encoder approaches~\cite{radford2021learning,jia2021scaling}.
Dual encoders learn a unified image-text feature space with separate encoders, thereby facilitating efficient nearest neighbor searches for recognition.
We use CLIP-L/14~\cite{radford2021learning}.

\noindent\textbf{Visual matching.}
We also experiment with pure visual matching baselines.
We use off-the-shelf CLIP-L/14 visual encoder and Entity-WebLI (55M) dataset as the memory.
We use $k=500$ for nearest neighbor search with majority voting as it obtains the best results on OVEN val set.

\noindent\textbf{Captioning.}
We compare to Git-Large~\cite{wang2022git} or PaLI~\cite{pali2022} image-to-text auto-regressive captioning models.

\noindent\textbf{\OURS-baselines: alternative code creation strategies.}
We compare \OURSf, \ie the best variant of \OURS, with several alternatives.
First, \OURSa refers to using atomic, completely unstructured codes, i.e. each code token $c^e_i$ is randomly drawn from $\llbracket 1, V \rrbracket^L$~\cite{tay2022transformer} .
Second, we consider two alternatives using semantically structured codes:
(i) \OURSh where we embed the entity names with a pretrained text encoder before applying hierarchical k-means clustering on the resulting embeddings~\cite{tay2022transformer}
and (ii) \OURSc where we create a code by tokenizing the entity name with $\Phi(.)$~\cite{hu2023open,de2020autoregressive}.
Details on the baselines are in Appendix Sec.~\ref{ap:sec:implem_baseline}.
\section{Experiments}
\label{sec:exp}

In this section, we detail our experimental setup, compare our method with state of the art and baselines, and finally present thorough analyses on code creation and pretraining.

\subsection{Experimental setting}
\label{sec:setting}

\noindent\textbf{OVEN dataset} consists of 6,063,945 different entities~\cite{hu2023open}.
We evaluate the models on the validation and test splits, by reporting the harmonic mean (HM) of top-1 accuracy scores between ``seen'' and ``unseen'' entities.
Seen are entities present in the OVEN training set.
Unseen entities are a subset of entities among the ones not present in the training set.
The models are evaluated on a total of 3192 entities (1721 for seen and 1471 for unseen) for validation and 15888 entities (8355 for seen and 7533 for unseen) for test.
We call the entities that the model is evaluated on by ``positive'' entities (\ie the union of the 3192 validation and 15888 test entities) and all other entities by ``negative'' entities.

\begin{table}[t]
\centering
\small
  \setlength{\tabcolsep}{0.8pt}
    \begin{tabular}{@{}l cccc c ccc @{}}
     \toprule
       & & \multicolumn{2}{c}{Pretraining} && \multicolumn{3}{c}{OVEN test}   \\
   \cmidrule{3-4} \cmidrule{6-8}
    Method & \#par.(B) & dataset & \#imgs && \colorbox{Gray}{HM} & \scriptsize{seen} & \scriptsize{unseen} \\
     \midrule
\multicolumn{5}{l}{\textit{Dual encoder approaches}} \\
  $\text{CLIP}_{\text{ViT-L14}}$ & 0.42 & OpenAI & 400M && \colorbox{Gray}{~5.2~} & 5.6 & 4.9\\
  $\text{CLIPfusion}_{\text{ViT-L14}}$ & 0.88 & OpenAI & 400M && \colorbox{Gray}{~8.4~} & 33.6 & 4.8 \\
  $\text{CLIP2CLIP}_{\text{ViT-L14}}$ & 0.86 & OpenAI & 400M &&\colorbox{Gray}{11.5} & 12.6 & 10.5 \\
    \midrule
  \multicolumn{5}{l}{\textit{Captioning approaches}} \\
  GiT-Large & 0.40 & WebLI & 100M && \colorbox{Gray}{~7.0~} & 17.6 & 4.3 \\
  PaLI-3B & 3 & WebLI & 1B && \colorbox{Gray}{~9.1~} & 19.1 & 6.0 \\
  PaLI-17B & 17 & WebLI & 1B &&  \colorbox{Gray}{16.0} & 28.3 & 11.2 \\
   \midrule
  \multicolumn{5}{l}{\textit{Generative entity recognition}} \\
  $\text{\OURSf}^\ddagger$ (Ours) & 0.40 & Entity-LAION & 41M && \colorbox{Gray}{20.9} & 29.1 & 16.3 \\
  \OURSf (Ours) & 0.40 & Entity-WebLI & 55M && \colorbox{Gray}{\textbf{22.7}} & 31.5 & 17.7 \\
  
      \bottomrule
 \end{tabular}
 \vspace{-0.2cm}
\caption{
      \textbf{Comparison with state-of-the-art} approaches on OVEN entity test split.
We report the harmonic mean (HM) of the seen and unseen splits (top-1 accuracy) after finetuning on OVEN training set.
Numbers are taken from~\cite{hu2023open} except methods based on GiT-Large which are run by us.
We indicate the total number of parameters of each model (``\# par.'') in billion and the pretraining dataset details.
$\ddagger$: use only publicly available data.
}
\vspace{-0.6cm}
    \label{tab:main_res}
\end{table}

\noindent\textbf{Pretraining and finetuning.}
Unless specified otherwise, we pretrain our models on the entity-WebLI dataset, which we create considering \textit{all} 6M entity names as described in Sec.~\ref{sec:training}.
After this entity-based pretraining, the models are finetuned on OVEN training set which consists only of the ``seen'' entities.
All implementation details are in Sec.~\ref{ap:sec:implem} in Appendix and code is released in the \textsc{scenic} library~\cite{dehghani2021scenic}.

\noindent\textbf{Preventing data leakage.}
We remove pretraining images from Entity-WebLI and Entity-LAION with a cosine similarity (with CLIP-L/14 visual features) above 0.95 with any of the OVEN test or val images.
We chose a 0.95 conservative threshold by looking at some examples: similarity 0.95 corresponds to conceptually similar images but clearly not duplicates (see Fig.~\ref{fig:leak} in Appendix).

\subsection{Comparison with the state of the art}
\label{sec:sota}
In Tab.~\ref{tab:main_res}, we compare the performance of \OURSf, our best \OURS variant, on the OVEN entity benchmark with previously published numbers after finetuning on the OVEN training set.
We see that our method outperforms previously proposed approaches by significant margins.
Notably, \OURSf improves over the captioning model PALI-17B by $+6.8$ top-1 HM test accuracy (a relative improvement of $43\%$) while using $42\times$ less parameters.

\subsection{Comparison with baselines}
\label{sec:baseline_res}
In Tab.~\ref{tab:main_baseline}, we compare \OURSf with the different baselines described in Sec.~\ref{sec:baselines}.
All baselines use exactly the \textit{same pretraining dataset} entity-based WebLI (55M) and model architectures of comparable sizes.

\noindent\textbf{Comparing \OURS to different paradigms.}
We see in Tab.~\ref{tab:main_baseline} that \OURS outperforms strong captioning, dual-encoder, visual matching and hierarchical classification baselines, affirming its advantage in tackling web-scale visual entity recognition.
Our superior performance compared to dual encoders aligns with previous works observing that CLIP struggles for fine-grained recognition~\cite{hu2023open,iscen2023retrieval}.
Due to query image and entity name similarities being captured only through a vector dot product, potentially fine-grained interactions are missed.
Also, \OURS offers significant advantages over dual encoders: its computational complexity is not a function of entity set size and it does not require to store entity dense embeddings.

\noindent\textbf{Different \OURS variants.}
In Tab.~\ref{tab:main_baseline}, we compare different variants of \OURS:
one variant using unstructured codes (\OURSa) and three variants using semantically-structured codes: \OURSc, \OURSh and \OURSf.
We observe that \OURSf is the best performing variant, both after entity-based pretraining and after finetuning on the OVEN seen entities.
Compared to \OURSc, \OURSf use codes that are more discriminative and compact, which improves the performance particularly for entities with long names (see Sec.~\ref{sec:captioning}).
Compared to \OURSa, \OURSf codes yield a semantic structure which is crucial for million-scale label-space as shown in Sec.~\ref{sec:atomic}.
\OURSh model also gets strong performance but relies on an off-the-shelf semantic text encoder which makes the approach more complex and costly compared to \OURSf.
\OURSh is a first step towards \textit{learning} codes and we hope future works will propose original and better code creation strategies~\cite{sun2023learning}.

\begin{table}[t]
\centering
\small
  \setlength{\tabcolsep}{1pt}
    \begin{tabular}{@{}l c ccc c ccc @{}}
     \toprule
     && \multicolumn{3}{c}{Entity-based pretraining} && \multicolumn{3}{c}{+ finetuning on seen}   \\
     \cmidrule{3-5} \cmidrule{7-9}
    Method && \colorbox{Gray}{HM} & \scriptsize{seen} & \scriptsize{unseen}  && \colorbox{Gray}{HM} & \scriptsize{seen} & \scriptsize{unseen} \\
     \midrule

\textit{Dual encoders} && \colorbox{Gray}{~9.2~} & 8.9 & 9.4 && \colorbox{Gray}{16.3} & 24.3 & 12.3 \\
\textit{Visual matching} && \colorbox{Gray}{16.2} & 15.5 & 17.1 && \colorbox{Gray}{16.4} & 15.7 & 17.2 \\
\textit{Captioning} && \colorbox{Gray}{13.2} & 13.1 & 13.3 && \colorbox{Gray}{16.8} & 25.9 & 12.5  \\
\textit{Hierarchic. classif.} && \colorbox{Gray}{14.7} & 14.8 & 14.6 && \colorbox{Gray}{21.8} & 29.6 & 17.2 \\

  \midrule
\multicolumn{8}{l}{\textit{Generative entity recognition}} \\
\OURSa && \colorbox{Gray}{15.9} & 15.3 & 16.7 && \colorbox{Gray}{20.1} & 26.2 & 16.3 \\
\OURSc && \colorbox{Gray}{14.3} & 16.5 & 12.6 && \colorbox{Gray}{20.7} & 26.8 & 16.9 \\
\OURSh && \colorbox{Gray}{15.8} & 15.5 & 16.0 && \colorbox{Gray}{21.0} & 25.2 & 17.9  \\
\OURSf && \colorbox{Gray}{\textbf{17.7}} & 18.3 & 17.2 && \colorbox{Gray}{\textbf{22.7}} & 31.5 & 17.7 \\
      \bottomrule
 \end{tabular}
  \vspace{-0.2cm}
\caption{
      \textbf{Baseline comparisons.}
All baselines use exactly the \textit{same pretraining dataset} Entity-WebLI (55M) and architectures of comparable number of parameters ($\sim 400$M).
All numbers are obtained with finetuning on seen split after entity-based pretraining.
We report the Harmonic Mean of top-1 accuracy on OVEN test.
}
\vspace{-0.4cm}
    \label{tab:main_baseline}
\end{table}

\subsection{Analysis and ablation study}
\label{sec:ablations}
In this section, unless specified otherwise, we report the accuracy on the OVEN validation set~\cite{hu2023open} evaluated after pretraining on Entity-WebLI (27M), \ie no OVEN finetuning.

\vspace{-0.3cm}
\subsubsection{Semantic versus atomic codes}
\begin{figure}[t]
\centering
\includegraphics[width=0.98\linewidth]{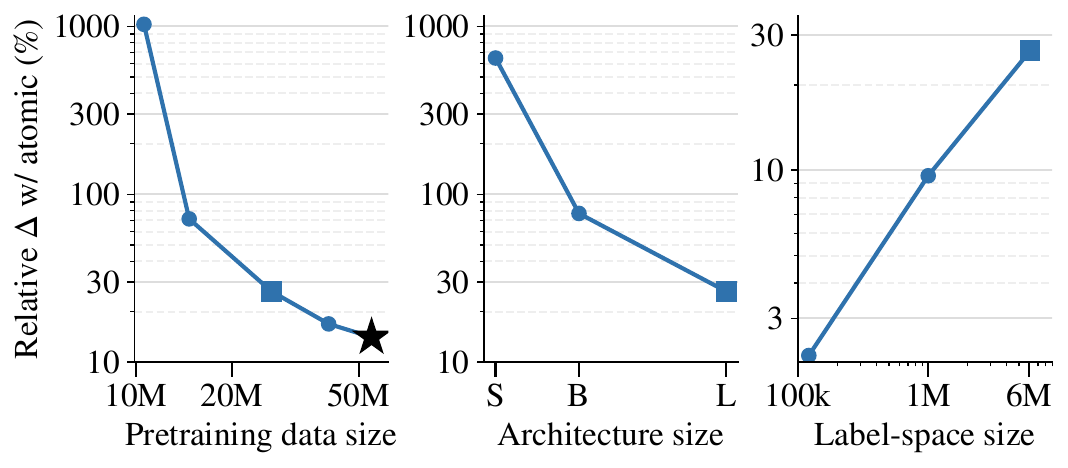}
 \vspace{-0.3cm}
\caption{\textbf{Semantic \textit{vs} atomic codes.}
We report the relative improvement in $\%$ of \OURSf compared to \OURSa in 3 scenarios: (i) limited pretraining data, (ii) limited model capacity and (iii) massive-scale label-space.
Plots share a common experiment shown by \textcolor{blue_plot}{$\mdblksquare$} which uses a pretraining dataset size of $27M$, Large model and 6M entity set.
The setting reported in Tab.~\ref{tab:main_baseline} is \small{{\FiveStar}}.}
\vspace{-0.5cm}
\label{fig:size_label_space}
\end{figure}

\vspace{-0.1cm}
\label{sec:atomic}
In Fig.~\ref{fig:size_label_space} (and Appendix Tab.~\ref{ap:tab:size_label_space}), we report the relative improvement of semantically-structured codes (\OURSf) compared to unstructured codes (\OURSa).
We vary pretraining data size, model capacity and label-space size.
A relative improvement of $100\%$ means that the performance of \OURSf doubles compared to \OURSa.

\noindent\textbf{Limited pretraining data.}
In Fig.~\ref{fig:size_label_space} (left), we see that semantic codes outperform atomic codes when the amount of data available for pretraining diminishes.
In fact, the results reported in Tab.~\ref{tab:main_baseline} corresponds to the most favorable scenario for \OURSa with 55M pretraining datapoints (represented by \small{\FiveStar} \normalsize in Fig.~\ref{fig:size_label_space}).
The relative improvement in this case is still of $14\%$ while it grows to more than $1000\%$ when the amount of data is reduced by $5 \times$.

\begin{figure*}[t]
\begin{minipage}{0.22\linewidth}
\centering
\includegraphics[width=1\linewidth]{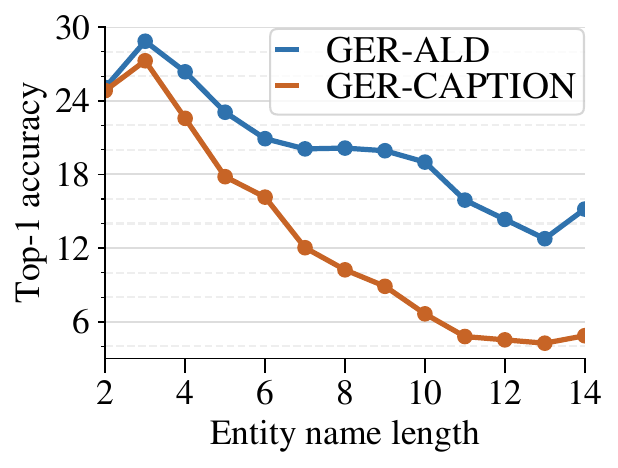}
\end{minipage}
\begin{minipage}{0.78\linewidth}
\centering
\includegraphics[width=1.015\linewidth]{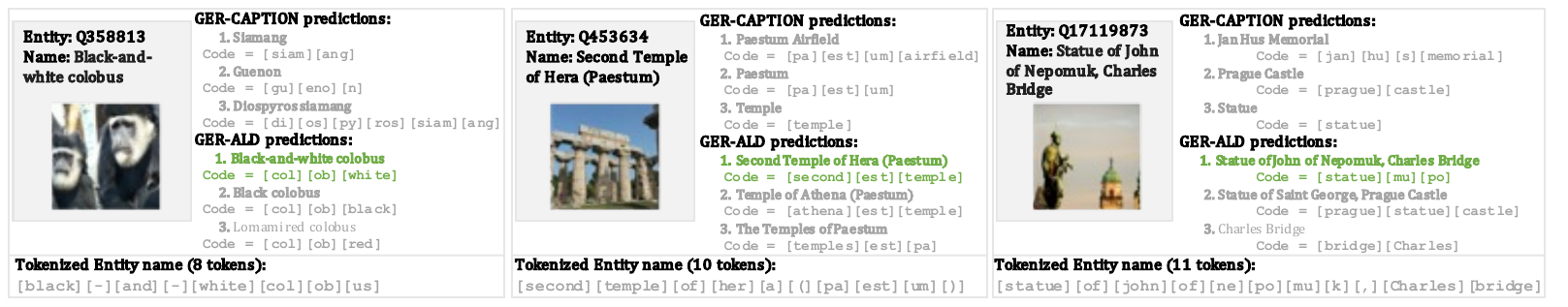}
\end{minipage}
\centering
\vspace{-0.4cm}
\caption{\textbf{Accuracy per entity name length for \OURSf versus \OURSc codes.}
(left): Accuracy averaged per entity name length.
(right): Qualitative examples of predictions for long entity names.
Code tokens are symbolized between brackets.
}
\vspace{-0.4cm}
\label{fig:vis_examples}
\end{figure*}

\begin{figure}[t]
\centering
\includegraphics[width=1\linewidth]{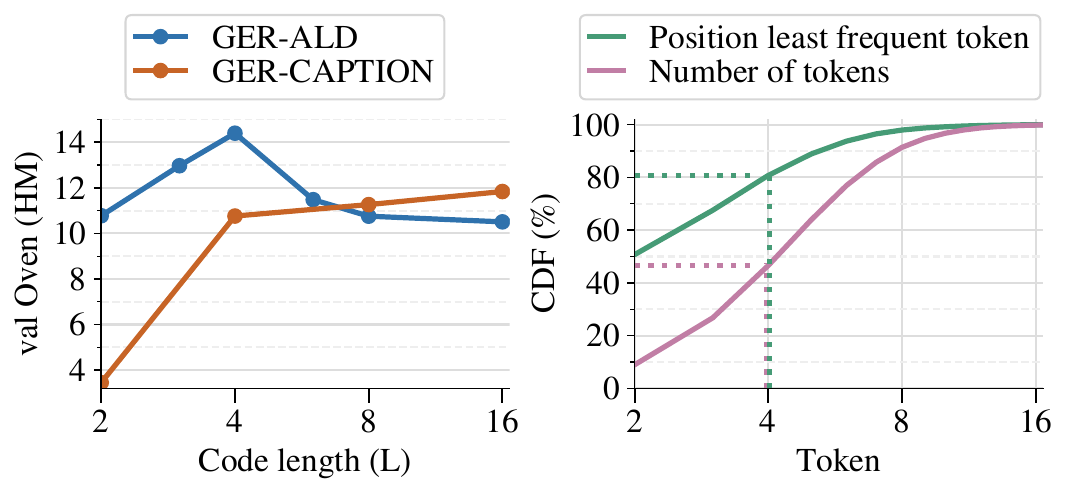}
\vspace{-0.8cm}
\caption{\textbf{\textsc{ald} versus captioning codes.}
(left): Effect of different code lengths for \OURSf and \OURSc codes.
(right): Cumulative distribution function (CDF) of (in green) the position of the least frequent token in the tokenized entity name and of (in pink) the length of tokenized entity name.
}
\vspace{-0.6cm}
\label{fig:code_length}
\end{figure}

\noindent\textbf{Limited model capacity.}
In Fig.~\ref{fig:size_label_space} (middle), we see that the model struggles to learn unstructured codes when its capacity is reduced.
When considering the small version of our model (114M parameters), the performance with atomic codes is very poor: $0.7$ top-1 accuracy.

\noindent\textbf{Web-scale label-space.}
In Fig.~\ref{fig:size_label_space} (right), we vary the number of entities for pretraining.
The ``positive'' entities (see Sec.~\ref{sec:setting}) are always included in the pretraining set and the amount of ``negative'' entities is increased, effectively acting as distractors.
First, we see in Fig.~\ref{fig:size_label_space} (right) that for relatively small-scale label-space ($\leq 100k$), the benefit of having semantic codes versus atomic is small.
In this regime we find that the model can memorize all the entities without the need for semantic structure between them.
This aligns with the findings of DSI~\cite{tay2022transformer}.
We evaluate \OURS further in small label-spaces in Sec.~\ref{sec:classif}.
However, we see that in million-scale label-space regime, semantic structure becomes important and significantly improves the performance compared to atomic codes: $+26\%$ relative improvement.

Overall, we find that \OURSa fail to learn unstructured codes when the amount of pretraining data or architecture capacity are reduced, or when the label-space increases to million-scale.
Unlike \OURSa, \OURSf succeed in these scenarios thanks to the semantic structure easing the learning.
Next, we analyze how \OURSf improves over another type of semantic codes: \OURSc codes.

\vspace{-0.2cm}
\subsubsection{\textsc{ald} versus captioning codes}
\vspace{-0.1cm}

\label{sec:captioning}
We analyze why unambiguous, language-based and discriminative codes (\OURSf) are more effective for entity recognition than directly decoding the entity name (\OURSc).
In Fig.~\ref{fig:code_length} (left), we report the performance of \OURSf and \OURSc when varying the length $L$ of the codes.
Fixing a code length $L$ to a caption corresponds to keeping only the first $L^{\text{th}}$ tokens of the entity name.
In Fig.~\ref{fig:code_length} (right), we report the cumulative distribution functions (CDF) of (i) the position within the tokenized entity name of the least frequent token among the entire corpus (as described in Sec.~\ref{sec:codes}) and (ii) the total number of tokens in the tokenized entity name ($L_e$ in the notations of Sec.~\ref{sec:codes})).

\noindent\textbf{Discriminative tokens versus number of tokens.}
We observe in Fig.~\ref{fig:code_length} (left) that the performance of \OURSc increases drastically from $L = 2$ to $L = 4$.
At the same time, we see in Fig.~\ref{fig:code_length} (right) that for $L = 4$, less than half of the entity names are considered in full while more than $80\%$ of the \OURSc codes contain the least frequent token of the entire tokenized name.
This hints that what is important for language-based codes is not to describe the full entity name but to include its most discriminative part.
We also observe that the performance of captioning increases only moderately from $L = 4$ to $L = 8$ even though the number of entities considered in full increases drastically from $46.6\%$ to $100\%$.
This confirms our intuition that decoding all the entity name tokens does not have a major impact on the performance as long as the most discriminative tokens are decoded.
Overall, these observations motivate the \textsc{ald} design of keeping only the most discriminative tokens, which is shown in Fig.~\ref{fig:code_length} to lead to improved performance compared to decoding the full tokenized entity name.

\noindent\textbf{Effect of code length for \OURSf.}
We see in Fig.~\ref{fig:code_length} (left) that the performance of \OURSf is the best for $L=4$.
With smaller code lengths, we need to resort to random tokens a lot to achieve unique codes (see Sec.~\ref{sec:codes}), which deters the performance.
For example at $L=2$, more than $10\%$ of the entities use a random code token while this percentage decreases to $0.5\%$ at $L=4$.
We also see that the performance of \OURSf decreases for code length above $L=4$, which hints that only the few most discriminative tokens are important while additional ones clutter the entity code.
Interestingly we also observe in Fig.~\ref{fig:code_length} (left) that when considering all the tokens,\OURSf performance is slightly below that of \OURSc.
This might seem surprising since the same amount of information is present in both cases.
However we find that when considering all the tokens, it is more difficult for the model to decode tokens ordered by frequencies than tokens ordered syntactically.

\noindent\textbf{Entities with long entity names.}
In Fig.~\ref{fig:vis_examples} (left), we report the accuracy per entity name length for both \OURSf and \OURSc finetuned models.
We see that the longer the entity name, the more \OURSf improves over captioning.
Longer entities tend to have more noise with key information further into the code.
We also show in Fig.~\ref{fig:vis_examples} qualitative examples of entities with long entity names (more in Fig.~\ref{ap:fig:vis} in Appendix).
In the left example, we see that \OURSf use the token combination [col][ob] to represent the semantic concept of colobus monkey species.
The last token is used to efficiently differentiate between sub-species of colobus.
This compact and discriminative way of encoding the entity allows \OURSf to successfully predict this entity whereas \OURSc fails to generate the entity tokenized name.

\begin{figure}[t]
\begin{minipage}{0.45\linewidth}
\centering
\small
\setlength{\tabcolsep}{7pt}
\begin{tabular}{@{}l c@{}}
  \toprule
   Selection strategy & HM \\
  \midrule
  Least frequent tokens  & 14.4 \\
  Most frequent tokens & 12.3 \\
  First tokens & 12.0 \\
  Random tokens & 11.3 \\
  \bottomrule
\end{tabular}
\end{minipage}\quad~~~
\begin{minipage}{0.45\linewidth}
\centering
\small
\setlength{\tabcolsep}{7pt}
\begin{tabular}{@{}l c @{}}
  \toprule
    Tokens order & HM \\
  \midrule
  Least frequent first  & 14.4 \\
  Syntax order & 14.4 \\
  Random order & 13.0 \\
  Least frequent last & 12.7 \\
  \bottomrule
 \end{tabular}
\end{minipage}
 \vspace{-0.2cm}
\captionof{table}{\textbf{Ablation study of \OURSf codes.}
(left) Word tokens selection.
(right) Tokens order.
All variants use $L=4$. Default is in top rows.
Non language-based \OURSa gets $11.4$ top-1.
}
\vspace{-0.4cm}
\label{fig:multi_ablat}
\end{figure}

\begin{figure}[t]
\begin{minipage}{0.4\linewidth}
\centering
\small
\setlength{\tabcolsep}{0.5pt}
\begin{tabular}{@{}lcc@{}}
  \toprule
   Dataset & Codes & HM \\
  \midrule
  WebLI & WebLI caption & 1.8~~~~~~~~~ \\
  Entity-WebLI \footnotesize{(55M)} & WebLI caption & 12.9 \textbf{\green{\tiny{(+11.1)}}} \\
  Entity-WebLI \footnotesize{(55M)} & Entity name & 14.8 \textbf{\green{\tiny{(+1.9)~}}} \\
  Entity-WebLI \footnotesize{(55M)} & \textsc{ALD} & 17.5 \textbf{\green{\tiny{(+2.7)~}}} \\
  \bottomrule
\end{tabular}
\end{minipage}\qquad~~~~~~~~~~~~~~~~~~~~~
\begin{minipage}{0.3\linewidth}
\centering
\includegraphics[width=1\linewidth]{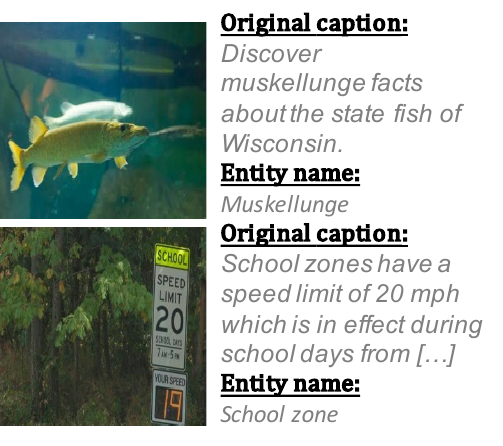}
\end{minipage}
\centering
\vspace{-0.2cm}
\caption{\textbf{Entity-based pretraining ablation.}
(left): Validation OVEN accuracy.
(right): Examples of original WebLI captions versus corresponding OVEN entity names.
}
\vspace{-0.4cm}
\label{fig:pretraining}
\end{figure}

\begin{figure}[t]
\begin{minipage}{0.55\linewidth}
\centering
\includegraphics[width=\linewidth]{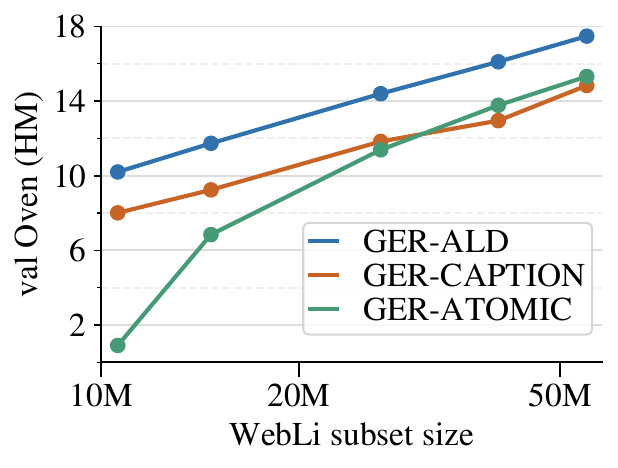}
\end{minipage}
  \begin{minipage}{0.4\linewidth}
	\caption{\textbf{Pretraining.}
We vary the size of the pretraining dataset by changing the amount of retrieved examples from WebLI for each OVEN entity (see Sec.~\ref{sec:training}).
	}
\label{fig:pretraining_size}
\end{minipage}
\vspace{-0.6cm}
\end{figure}

 \vspace{-0.4cm}
\subsubsection{Creating codes with \textsc{ald}}
 \vspace{-0.2cm}
 
\noindent\textbf{Least frequent tokens.}
In Tab.~\ref{fig:multi_ablat} (left), we validate our choice of selecting the least frequent tokens by evaluating 3 alternatives: random choice, most frequent tokens and first-appearing tokens in tokenized entity name.
We see that these alternative strategies hurt the performance significantly.
Qualitative examples in Appendix Fig.~\ref{ap:fig:ablation_sdu} show that the kept tokens are less semantic and discriminative compared to \OURSf strategy of keeping the least frequent tokens.
Note that all these variants are at least as good as \OURSa ($11.4$ top-1) which is not based on language at all.

\noindent\textbf{Decoding order.}
In Tab.~\ref{fig:multi_ablat} (right), we vary the order of the first $L-1$ tokens in \OURSf codes.
Instead of decoding tokens from least to most frequent, we evaluate most to least frequent, syntax order and random order.
Note that the selected tokens are the same in all variants, only their order changes.
We see that both ``least frequent first'' and ``syntax'' orders achieve the best of performance.

 \vspace{-0.2cm}
\subsubsection{Entity-based pretraining}

\noindent\textbf{Entity-based pretraining.}
In Fig.~\ref{fig:pretraining}, we analyze why our entity-based pretraining improves over the standard captioning pretraining of PaLI or GiT models.
First, we see that our method of selecting WebLI data relevant to OVEN entities drastically improves the performance (+11.1 in Fig.~\ref{fig:pretraining} (left)).
This is because, by design, we select image-text pairs from WebLI that have captions similar to OVEN entity names.
Hence, this data is directly relevant for the OVEN entity recognition benchmark.
Second, we see that replacing the original WebLI caption with its corresponding entity name from OVEN leads to superior performance (+1.9).
We see in the qualitative examples of Fig.~\ref{fig:pretraining} (right) that original captions contain a lot of descriptive information not directly relevant to the entity.
Lastly, we confirm that using \OURSf codes is better (+2.7) than tokenized entity name.

\noindent\textbf{Dataset size.}
In Fig.~\ref{fig:pretraining_size}, we evaluate the effect of the pretraining dataset size for \OURS models.
We control the dataset size by varying the amount of retrieved examples from WebLI for each of the OVEN entities (see Sec.~\ref{sec:training}).
We see in Fig.~\ref{fig:pretraining_size} that \OURSf, \OURSc and \OURSa benefit greatly from more data and do not seem to have reached saturation yet.
As analyzed in Sec.~\ref{sec:atomic}, \OURSa fails when the amount of pretraining data decreases.

\subsection{Link with classification}
\label{sec:classif}

\begin{table}
\centering
\small
\setlength{\tabcolsep}{2pt}
\begin{tabular}{@{}l c c@{}}
        \toprule
                Method                                                & ImageNet-LT    & WebVision     \\
        \midrule
        Classif. MLP                                         & 74.3              & 80.9                  \\
         \OURSa $L = 2$                                            & 80.8              & 84.7                  \\
        \OURSf $L = 2$                                                 & 80.9              & \textbf{84.8}                  \\
        \OURSa $L = 1$ ($\sim$ Classif. MAP)                                            & \textbf{81.0}              & \textbf{84.8}             \\
        \midrule
        \multicolumn{3}{l}{\textit{Previously published numbers}}                                              \\
           NCR~\cite{iscen2022learning}                            & --                & 76.8                  \\
        CurrNet~\cite{guo2018curriculumnet}                     & --                & 79.3                 \\
        PEL~\cite{shi2023parameter}                             & 78.3              & --                     \\
        MAM~\cite{iscen2023improving}$\dagger$                  & 82.3              & 83.6                   \\
     
        \bottomrule
        \end{tabular}
         \vspace{-0.2cm}
  \caption{\textbf{Evaluation of classification models and \OURS on small-scale label-spaces.}
$\dagger$ indicates the use of additional data.
  \label{tab:classification}}
  \vspace{-0.6cm}
\end{table}

A typical way of tackling visual entity recognition is by training a classifier into the number of entities~\cite{russakovsky2015imagenet}.
This is not a viable solution for web-scale problems such as OVEN where a single fully-connected layer for a $6$M classes has an enormous parameter count of $4.6$B. 
In this section, we evaluate \OURS in cases where learning a classification model is a feasible choice (smaller number of classes).
Classification can be cast in our \OURS framework simply by setting $L=1$ and $V = |\mathcal{E}| = $ number of classes (see Sec.~\ref{sec:problem}), making it a special case of atomic codes with $L=1$.
Since the decoder decodes a single token, it is equivalent to a multi-layer Multihead Attention Pooling (MAP) head~\cite{zhai2022scaling,lee2019set}.
In Tab.~\ref{tab:classification}, we consider two challenging classification datasets: long-tailed ImageNet-LT~\cite{OLTR} and noisy Webvision~\cite{li2017webvision}.
We evaluate \OURS-\{\textsc{ald}, \textsc{atomic}\} and a classification baseline using multi-layer perceptron (MLP) on averaged-pooled patch tokens.
Implementation details are in Sec~\ref{ap:sec:classif} in Appendix.

We see in Tab.~\ref{tab:classification} that using \OURSa instead of standard MLP improves significantly the performance of the classification model (74.3 versus 81.0 for ImageNet-LT).
We also observe that \OURSa and \OURSf have comparable performance in this relatively small label-space regime (1k classes).
As a matter of fact, this achieves state-of-the-art accuracy for both datasets (when no additional external data is used).
This shows that \OURS framework not only excels for large-scale scenarios, but also works well in datasets with smaller number of visual entities, making \OURS a general framework for visual entity recognition.
\section{Conclusion}
In this work, we propose a novel generative framework for web-scale visual entity recognition.
We represent each entity by a compact, discriminative and semantic code that a generative auto-regressive model learns to decode.
In future work, we will explore ways of creating better entity codes by leveraging additional information: either from the Wikipedia page such as the description of the entity and its attached image or also by using external tools.
\vspace{-0.4cm}
\paragraph{Acknowledgement.}
We thank Xingyi Zhou, Ziniu Hu and Armand Joulin, as well as our teammates for their precious help, support and discussions around this project.
{
    \small
    \bibliographystyle{ieeenat_fullname}
    \bibliography{main}

\begin{thebibliography}{50}
\providecommand{\natexlab}[1]{#1}
\providecommand{\url}[1]{\texttt{#1}}
\expandafter\ifx\csname urlstyle\endcsname\relax
  \providecommand{\doi}[1]{doi: #1}\else
  \providecommand{\doi}{doi: \begingroup \urlstyle{rm}\Url}\fi

\bibitem[Agrawal et~al.(2013)Agrawal, Gupta, Prabhu, and
  Varma]{agrawal2013multi}
Rahul Agrawal, Archit Gupta, Yashoteja Prabhu, and Manik Varma.
\newblock Multi-label learning with millions of labels: Recommending advertiser
  bid phrases for web pages.
\newblock In \emph{Proceedings of the 22nd international conference on World
  Wide Web}, pages 13--24, 2013.

\bibitem[Alayrac et~al.(2022)Alayrac, Donahue, Luc, Miech, Barr, Hasson, Lenc,
  Mensch, Millican, Reynolds, et~al.]{alayrac2022flamingo}
Jean-Baptiste Alayrac, Jeff Donahue, Pauline Luc, Antoine Miech, Iain Barr,
  Yana Hasson, Karel Lenc, Arthur Mensch, Katherine Millican, Malcolm Reynolds,
  et~al.
\newblock Flamingo: a visual language model for few-shot learning.
\newblock \emph{Advances in Neural Information Processing Systems},
  35:\penalty0 23716--23736, 2022.

\bibitem[Bengio et~al.(2019)Bengio, Dembczynski, Joachims, Kloft, and
  Varma]{bengio_et_al:DagRep.8.7.62}
Samy Bengio, Krzysztof Dembczynski, Thorsten Joachims, Marius Kloft, and Manik
  Varma.
\newblock {Extreme Classification (Dagstuhl Seminar 18291)}.
\newblock \emph{Dagstuhl Reports}, 2019.

\bibitem[Bossard et~al.(2014)Bossard, Guillaumin, and
  Van~Gool]{bossard2014food}
Lukas Bossard, Matthieu Guillaumin, and Luc Van~Gool.
\newblock Food-101--mining discriminative components with random forests.
\newblock In \emph{ECCV}, 2014.

\bibitem[Chen et~al.(2023)Chen, Wang, Changpinyo, Piergiovanni, Padlewski,
  Salz, Goodman, Grycner, Mustafa, Beyer, et~al.]{pali2022}
Xi Chen, Xiao Wang, Soravit Changpinyo, AJ Piergiovanni, Piotr Padlewski,
  Daniel Salz, Sebastian Goodman, Adam Grycner, Basil Mustafa, Lucas Beyer,
  et~al.
\newblock Pali: A jointly-scaled multilingual language-image model.
\newblock \emph{ICLR}, 2023.

\bibitem[De~Cao et~al.(2020)De~Cao, Izacard, Riedel, and
  Petroni]{de2020autoregressive}
Nicola De~Cao, Gautier Izacard, Sebastian Riedel, and Fabio Petroni.
\newblock Autoregressive entity retrieval.
\newblock \emph{arXiv preprint arXiv:2010.00904}, 2020.

\bibitem[Dehghani et~al.(2021)Dehghani, Gritsenko, Arnab, Minderer, and
  Tay]{dehghani2021scenic}
Mostafa Dehghani, Alexey Gritsenko, Anurag Arnab, Matthias Minderer, and Yi
  Tay.
\newblock {Scenic}: A {JAX} library for computer vision research and beyond.
\newblock \emph{arXiv preprint arXiv:2110.11403}, 2021.

\bibitem[Everingham et~al.(2010)Everingham, Van~Gool, Williams, Winn, and
  Zisserman]{everingham2010pascal}
Mark Everingham, Luc Van~Gool, Christopher~KI Williams, John Winn, and Andrew
  Zisserman.
\newblock The pascal visual object classes (voc) challenge.
\newblock \emph{IJCV}, 88, 2010.

\bibitem[Fei-Fei et~al.(2004)Fei-Fei, Fergus, and Perona]{fei2004learning}
Li Fei-Fei, Rob Fergus, and Pietro Perona.
\newblock Learning generative visual models from few training examples: An
  incremental bayesian approach tested on 101 object categories.
\newblock In \emph{CVPR}, 2004.

\bibitem[Guo et~al.(2018)Guo, Huang, Zhang, Zhuang, Dong, Scott, and
  Huang]{guo2018curriculumnet}
Sheng Guo, Weilin Huang, Haozhi Zhang, Chenfan Zhuang, Dengke Dong, Matthew~R
  Scott, and Dinglong Huang.
\newblock {CurriculumNet}: Weakly supervised learning from large-scale web
  images.
\newblock In \emph{ECCV}, pages 135--150, 2018.

\bibitem[Gutmann and Hyv{\"a}rinen(2010)]{gutmann2010noise}
Michael Gutmann and Aapo Hyv{\"a}rinen.
\newblock Noise-contrastive estimation: A new estimation principle for
  unnormalized statistical models.
\newblock In \emph{Proceedings of the thirteenth international conference on
  artificial intelligence and statistics}. JMLR Workshop and Conference
  Proceedings, 2010.

\bibitem[Hu et~al.(2023)Hu, Luan, Chen, Khandelwal, Joshi, Lee, Toutanova, and
  Chang]{hu2023open}
Hexiang Hu, Yi Luan, Yang Chen, Urvashi Khandelwal, Mandar Joshi, Kenton Lee,
  Kristina Toutanova, and Ming-Wei Chang.
\newblock Open-domain visual entity recognition: Towards recognizing millions
  of wikipedia entities.
\newblock \emph{ICCV}, 2023.

\bibitem[Iscen et~al.(2022)Iscen, Valmadre, Arnab, and
  Schmid]{iscen2022learning}
Ahmet Iscen, Jack Valmadre, Anurag Arnab, and Cordelia Schmid.
\newblock Learning with neighbor consistency for noisy labels.
\newblock In \emph{CVPR}, 2022.

\bibitem[Iscen et~al.(2023)Iscen, Fathi, and Schmid]{iscen2023improving}
Ahmet Iscen, Alireza Fathi, and Cordelia Schmid.
\newblock Improving image recognition by retrieving from web-scale image-text
  data.
\newblock \emph{CVPR}, 2023.

\bibitem[Iscen et~al.(2024)Iscen, Caron, Fathi, and Schmid]{iscen2023retrieval}
Ahmet Iscen, Mathilde Caron, Alireza Fathi, and Cordelia Schmid.
\newblock Retrieval-enhanced contrastive vision-text models.
\newblock \emph{ICLR}, 2024.

\bibitem[Jia et~al.(2021)Jia, Yang, Xia, Chen, Parekh, Pham, Le, Sung, Li, and
  Duerig]{jia2021scaling}
Chao Jia, Yinfei Yang, Ye Xia, Yi-Ting Chen, Zarana Parekh, Hieu Pham, Quoc Le,
  Yun-Hsuan Sung, Zhen Li, and Tom Duerig.
\newblock Scaling up visual and vision-language representation learning with
  noisy text supervision.
\newblock In \emph{ICML}, 2021.

\bibitem[Khosla et~al.(2011)Khosla, Jayadevaprakash, Yao, and
  Fei-Fei]{KhoslaYaoJayadevaprakashFeiFei_FGVC2011}
Aditya Khosla, Nityananda Jayadevaprakash, Bangpeng Yao, and Li Fei-Fei.
\newblock Novel dataset for fine-grained image categorization.
\newblock In \emph{First Workshop on Fine-Grained Visual Categorization, CVPR},
  2011.

\bibitem[Krause et~al.(2013)Krause, Stark, Deng, and Fei-Fei]{krause20133d}
Jonathan Krause, Michael Stark, Jia Deng, and Li Fei-Fei.
\newblock 3d object representations for fine-grained categorization.
\newblock In \emph{ICCV}, 2013.

\bibitem[Kudo(2018)]{kudo2018subword}
Taku Kudo.
\newblock Subword regularization: Improving neural network translation models
  with multiple subword candidates.
\newblock \emph{arXiv preprint arXiv:1804.10959}, 2018.

\bibitem[Kudo and Richardson(2018)]{kudo2018sentencepiece}
Taku Kudo and John Richardson.
\newblock Sentencepiece: A simple and language independent subword tokenizer
  and detokenizer for neural text processing.
\newblock \emph{arXiv preprint arXiv:1808.06226}, 2018.

\bibitem[Lee et~al.(2019)Lee, Lee, Kim, Kosiorek, Choi, and Teh]{lee2019set}
Juho Lee, Yoonho Lee, Jungtaek Kim, Adam Kosiorek, Seungjin Choi, and Yee~Whye
  Teh.
\newblock Set transformer: A framework for attention-based
  permutation-invariant neural networks.
\newblock In \emph{International conference on machine learning}, 2019.

\bibitem[Li et~al.(2017)Li, Wang, Li, Agustsson, and Van~Gool]{li2017webvision}
Wen Li, Limin Wang, Wei Li, Eirikur Agustsson, and Luc Van~Gool.
\newblock Webvision database: Visual learning and understanding from web data.
\newblock \emph{arXiv preprint arXiv:1708.02862}, 2017.

\bibitem[Liu et~al.(2023)Liu, Son, Yang, Liu, Gao, Lee, and
  Li]{liu2023learning}
Haotian Liu, Kilho Son, Jianwei Yang, Ce Liu, Jianfeng Gao, Yong~Jae Lee, and
  Chunyuan Li.
\newblock Learning customized visual models with retrieval-augmented knowledge.
\newblock In \emph{CVPR}, 2023.

\bibitem[Liu et~al.(2019)Liu, Miao, Zhan, Wang, Gong, and Yu]{OLTR}
Ziwei Liu, Zhongqi Miao, Xiaohang Zhan, Jiayun Wang, Boqing Gong, and Stella~X.
  Yu.
\newblock Large-scale long-tailed recognition in an open world.
\newblock In \emph{CVPR}, 2019.

\bibitem[Mehta et~al.(2022)Mehta, Gupta, Tay, Dehghani, Tran, Rao, Najork,
  Strubell, and Metzler]{mehta2022dsi++}
Sanket~Vaibhav Mehta, Jai Gupta, Yi Tay, Mostafa Dehghani, Vinh~Q Tran, Jinfeng
  Rao, Marc Najork, Emma Strubell, and Donald Metzler.
\newblock Dsi++: Updating transformer memory with new documents.
\newblock \emph{arXiv preprint arXiv:2212.09744}, 2022.

\bibitem[Mittal et~al.(2022)Mittal, Dahiya, Malani, Ramaswamy, Kuruvilla,
  Ajmera, Chang, Agarwal, Kar, and Varma]{mittal2022multi}
Anshul Mittal, Kunal Dahiya, Shreya Malani, Janani Ramaswamy, Seba Kuruvilla,
  Jitendra Ajmera, Keng-hao Chang, Sumeet Agarwal, Purushottam Kar, and Manik
  Varma.
\newblock Multi-modal extreme classification.
\newblock In \emph{CVPR}, 2022.

\bibitem[M{\"u}ller et~al.(2019)M{\"u}ller, Kornblith, and
  Hinton]{muller2019does}
Rafael M{\"u}ller, Simon Kornblith, and Geoffrey~E Hinton.
\newblock When does label smoothing help?
\newblock \emph{Advances in neural information processing systems}, 32, 2019.

\bibitem[Ni et~al.(2021)Ni, {\'A}brego, Constant, Ma, Hall, Cer, and
  Yang]{ni2021sentence}
Jianmo Ni, Gustavo~Hern{\'a}ndez {\'A}brego, Noah Constant, Ji Ma, Keith~B
  Hall, Daniel Cer, and Yinfei Yang.
\newblock Sentence-t5: Scalable sentence encoders from pre-trained text-to-text
  models.
\newblock \emph{arXiv preprint arXiv:2108.08877}, 2021.

\bibitem[Oord et~al.(2018)Oord, Li, and Vinyals]{oord2018representation}
Aaron van~den Oord, Yazhe Li, and Oriol Vinyals.
\newblock Representation learning with contrastive predictive coding.
\newblock \emph{arXiv preprint arXiv:1807.03748}, 2018.

\bibitem[OpenAI(2023)]{gpt4}
OpenAI.
\newblock {GPT-4} technical report.
\newblock \emph{arXiv preprint arXiv:2303.08774}, 2023.

\bibitem[Pradeep et~al.(2023)Pradeep, Hui, Gupta, Lelkes, Zhuang, Lin, Metzler,
  and Tran]{pradeep2023does}
Ronak Pradeep, Kai Hui, Jai Gupta, Adam~D Lelkes, Honglei Zhuang, Jimmy Lin,
  Donald Metzler, and Vinh~Q Tran.
\newblock How does generative retrieval scale to millions of passages?
\newblock \emph{arXiv preprint arXiv:2305.11841}, 2023.

\bibitem[Radford et~al.(2021)Radford, Kim, Hallacy, Ramesh, Goh, Agarwal,
  Sastry, Askell, Mishkin, Clark, et~al.]{radford2021learning}
Alec Radford, Jong~Wook Kim, Chris Hallacy, Aditya Ramesh, Gabriel Goh,
  Sandhini Agarwal, Girish Sastry, Amanda Askell, Pamela Mishkin, Jack Clark,
  et~al.
\newblock Learning transferable visual models from natural language
  supervision.
\newblock In \emph{ICML}, 2021.

\bibitem[Rajput et~al.(2023)Rajput, Mehta, Singh, Keshavan, Vu, Heldt, Hong,
  Tay, Tran, Samost, et~al.]{rajput2023recommender}
Shashank Rajput, Nikhil Mehta, Anima Singh, Raghunandan~H Keshavan, Trung Vu,
  Lukasz Heldt, Lichan Hong, Yi Tay, Vinh~Q Tran, Jonah Samost, et~al.
\newblock Recommender systems with generative retrieval.
\newblock \emph{arXiv preprint arXiv:2305.05065}, 2023.

\bibitem[Robertson et~al.(2009)Robertson, Zaragoza,
  et~al.]{robertson2009probabilistic}
Stephen Robertson, Hugo Zaragoza, et~al.
\newblock The probabilistic relevance framework: Bm25 and beyond.
\newblock \emph{Foundations and Trends{\textregistered} in Information
  Retrieval}, 2009.

\bibitem[Russakovsky et~al.(2015)Russakovsky, Deng, Su, Krause, Satheesh, Ma,
  Huang, Karpathy, Khosla, Bernstein, et~al.]{russakovsky2015imagenet}
Olga Russakovsky, Jia Deng, Hao Su, Jonathan Krause, Sanjeev Satheesh, Sean Ma,
  Zhiheng Huang, Andrej Karpathy, Aditya Khosla, Michael Bernstein, et~al.
\newblock Imagenet large scale visual recognition challenge.
\newblock \emph{International journal of computer vision}, 2015.

\bibitem[Schroff et~al.(2015)Schroff, Kalenichenko, and
  Philbin]{schroff2015facenet}
Florian Schroff, Dmitry Kalenichenko, and James Philbin.
\newblock Facenet: A unified embedding for face recognition and clustering.
\newblock In \emph{CVPR}, 2015.

\bibitem[Schuhmann et~al.(2021)Schuhmann, Vencu, Beaumont, Kaczmarczyk, Mullis,
  Katta, Coombes, Jitsev, and Komatsuzaki]{schuhmann2021laion}
Christoph Schuhmann, Richard Vencu, Romain Beaumont, Robert Kaczmarczyk,
  Clayton Mullis, Aarush Katta, Theo Coombes, Jenia Jitsev, and Aran
  Komatsuzaki.
\newblock Laion-400m: Open dataset of clip-filtered 400 million image-text
  pairs.
\newblock \emph{arXiv preprint arXiv:2111.02114}, 2021.

\bibitem[Schuhmann et~al.(2022)Schuhmann, Beaumont, Vencu, Gordon, Wightman,
  Cherti, Coombes, Katta, Mullis, Wortsman, et~al.]{schuhmann2022laion}
Christoph Schuhmann, Romain Beaumont, Richard Vencu, Cade Gordon, Ross
  Wightman, Mehdi Cherti, Theo Coombes, Aarush Katta, Clayton Mullis, Mitchell
  Wortsman, et~al.
\newblock Laion-5b: An open large-scale dataset for training next generation
  image-text models.
\newblock \emph{arXiv preprint arXiv:2210.08402}, 2022.

\bibitem[Sennrich et~al.(2015)Sennrich, Haddow, and Birch]{sennrich2015neural}
Rico Sennrich, Barry Haddow, and Alexandra Birch.
\newblock Neural machine translation of rare words with subword units.
\newblock \emph{arXiv preprint arXiv:1508.07909}, 2015.

\bibitem[Shi et~al.(2023)Shi, Wei, Zhou, Han, Shao, and Li]{shi2023parameter}
Jiang-Xin Shi, Tong Wei, Zhi Zhou, Xin-Yan Han, Jie-Jing Shao, and Yu-Feng Li.
\newblock Parameter-efficient long-tailed recognition.
\newblock \emph{arXiv preprint arXiv:2309.10019}, 2023.

\bibitem[Sun et~al.(2023)Sun, Yan, Chen, Wang, Zhu, Ren, Chen, Yin, Rijke, and
  Ren]{sun2023learning}
Weiwei Sun, Lingyong Yan, Zheng Chen, Shuaiqiang Wang, Haichao Zhu, Pengjie
  Ren, Zhumin Chen, Dawei Yin, Maarten Rijke, and Zhaochun Ren.
\newblock Learning to tokenize for generative retrieval.
\newblock \emph{NeurIPS}, 2023.

\bibitem[Tay et~al.(2022)Tay, Tran, Dehghani, Ni, Bahri, Mehta, Qin, Hui, Zhao,
  Gupta, et~al.]{tay2022transformer}
Yi Tay, Vinh Tran, Mostafa Dehghani, Jianmo Ni, Dara Bahri, Harsh Mehta, Zhen
  Qin, Kai Hui, Zhe Zhao, Jai Gupta, et~al.
\newblock Transformer memory as a differentiable search index.
\newblock \emph{Advances in Neural Information Processing Systems}, 2022.

\bibitem[Van~Horn et~al.(2018)Van~Horn, Mac~Aodha, Song, Cui, Sun, Shepard,
  Adam, Perona, and Belongie]{van2018inaturalist}
Grant Van~Horn, Oisin Mac~Aodha, Yang Song, Yin Cui, Chen Sun, Alex Shepard,
  Hartwig Adam, Pietro Perona, and Serge Belongie.
\newblock The inaturalist species classification and detection dataset.
\newblock In \emph{CVPR}, 2018.

\bibitem[Wah et~al.(2011)Wah, Branson, Welinder, Perona, and
  Belongie]{wah2011caltech}
Catherine Wah, Steve Branson, Peter Welinder, Pietro Perona, and Serge
  Belongie.
\newblock The caltech-ucsd birds-200-2011 dataset.
\newblock 2011.

\bibitem[Wang et~al.(2022{\natexlab{a}})Wang, Yang, Hu, Li, Lin, Gan, Liu, Liu,
  and Wang]{wang2022git}
Jianfeng Wang, Zhengyuan Yang, Xiaowei Hu, Linjie Li, Kevin Lin, Zhe Gan,
  Zicheng Liu, Ce Liu, and Lijuan Wang.
\newblock Git: A generative image-to-text transformer for vision and language.
\newblock \emph{arXiv preprint arXiv:2205.14100}, 2022{\natexlab{a}}.

\bibitem[Wang et~al.(2022{\natexlab{b}})Wang, Hou, Wang, Miao, Wu, Chen, Xia,
  Chi, Zhao, Liu, et~al.]{wang2022neural}
Yujing Wang, Yingyan Hou, Haonan Wang, Ziming Miao, Shibin Wu, Qi Chen, Yuqing
  Xia, Chengmin Chi, Guoshuai Zhao, Zheng Liu, et~al.
\newblock A neural corpus indexer for document retrieval.
\newblock \emph{NeurIPS}, 2022{\natexlab{b}}.

\bibitem[Weyand et~al.(2020)Weyand, Araujo, Cao, and Sim]{weyand2020google}
Tobias Weyand, Andre Araujo, Bingyi Cao, and Jack Sim.
\newblock Google landmarks dataset v2-a large-scale benchmark for
  instance-level recognition and retrieval.
\newblock In \emph{CVPR}, 2020.

\bibitem[Zhai et~al.(2022)Zhai, Kolesnikov, Houlsby, and
  Beyer]{zhai2022scaling}
Xiaohua Zhai, Alexander Kolesnikov, Neil Houlsby, and Lucas Beyer.
\newblock Scaling vision transformers.
\newblock In \emph{CVPR}, 2022.

\bibitem[Zhang et~al.(2023)Zhang, Zhang, Chen, Wang, Chen, Xie, Sun, Deng,
  Zhang, Yang, et~al.]{zhang2023irgen}
Yidan Zhang, Ting Zhang, Dong Chen, Yujing Wang, Qi Chen, Xing Xie, Hao Sun,
  Weiwei Deng, Qi Zhang, Fan Yang, et~al.
\newblock Irgen: Generative modeling for image retrieval.
\newblock \emph{arXiv preprint arXiv:2303.10126}, 2023.

\bibitem[Zhu et~al.(2022)Zhu, Huang, Deng, Ye, Huang, Chen, Zhu, Yang, Du, Lu,
  et~al.]{zhu2022webface260m}
Zheng Zhu, Guan Huang, Jiankang Deng, Yun Ye, Junjie Huang, Xinze Chen, Jiagang
  Zhu, Tian Yang, Dalong Du, Jiwen Lu, et~al.
\newblock Webface260m: A benchmark for million-scale deep face recognition.
\newblock \emph{IEEE Transactions on Pattern Analysis and Machine
  Intelligence}, 2022.

\end{thebibliography}
}
\clearpage
\maketitlesupplementary

\begin{table}[b]
\centering
\small
  \setlength{\tabcolsep}{3pt}
    \begin{tabular}{@{}l c ccc c ccc @{}}
     \toprule
     && \multicolumn{3}{c}{Zero-shot} && \multicolumn{3}{c}{Finetuned on OVEN}   \\
     \cmidrule{3-5} \cmidrule{7-9}
    Method && \colorbox{Gray}{HM} & \scriptsize{Seen} & \scriptsize{Unseen}  && \colorbox{Gray}{HM} & \scriptsize{Seen} & \scriptsize{Unseen} \\
     \midrule
PaLI-17B~\cite{pali2022} && \colorbox{Gray}{~1.8~} & 4.4 & 1.2 && \colorbox{Gray}{16.0} & 28.3 & 11.2 \\
GiT-Large~\cite{wang2022git} && \colorbox{Gray}{~1.7~} & 4.1 & 1.2 && \colorbox{Gray}{~7.0~} & 17.6 & 4.3 \\
      \bottomrule
 \end{tabular}
\caption{
      \textbf{Transferring captioning models to OVEN.}
We report the harmonic mean (HM) of top-1 accuracy on the seen and unseen test splits for two captioning models: PALI-17B~\cite{pali2022} and GiT-Large~\cite{wang2022git}.
Numbers from GiT-Large are run by us.
Note that GiT-Large has $42\times$ less parameters thank PALI-17B.
}
    \label{ap:tab:captioning_zeroshot}
\end{table}

\section{Implementation Details}
\label{ap:sec:implem}
We use the Large version of GIT~\cite{wang2022git} with a pretrained visual encoder and a decoder randomly initialized.
The visual encoder is pre-trained with GIT trained for captioning on WebLI dataset~\cite{wang2022git,pali2022}.

\subsection{Entity-based pre-training}
We use batch size of 4096, learning rate of $1e-5$ for the visual encoder and $1\text{e}^{-4}$ for the decoder, label smoothing of $0.3$ and no weight decay.
We use standard inception crop data augmentation for the images.
By default and unless specified otherwise, we use code length $L=4$ (see Fig.~\ref{fig:code_length}).
Note that we only evaluate codes with $L > 1$ on OVEN, as the only way to ensure unique codes with $L = 1$ is to set $V = |\mathcal{E}|$.
This is equivalent to the classification scenario and is not feasible for the million-scale label-space of OVEN.
We evaluate $L=1$ in Sec.~\ref{sec:classif} for datasets with a smaller label-space of 1k entities: namely ImageNet-LT~\cite{OLTR} and Webvision~\cite{li2017webvision}.
Unless specified otherwise our models for the main results (i.e. in Sec.~\ref{sec:sota} and Sec.~\ref{sec:baseline_res}) are trained on Entity-WebLI with 55M images ($k=100$) during 600k steps while models for ablations are trained on Entity-WebLI with 27M images ($k=20$) for 200k steps.

\begin{figure}[ht]
\small
 \begin{minipage}{0.4\linewidth}
    \centering
    \includegraphics[width=0.88\linewidth]{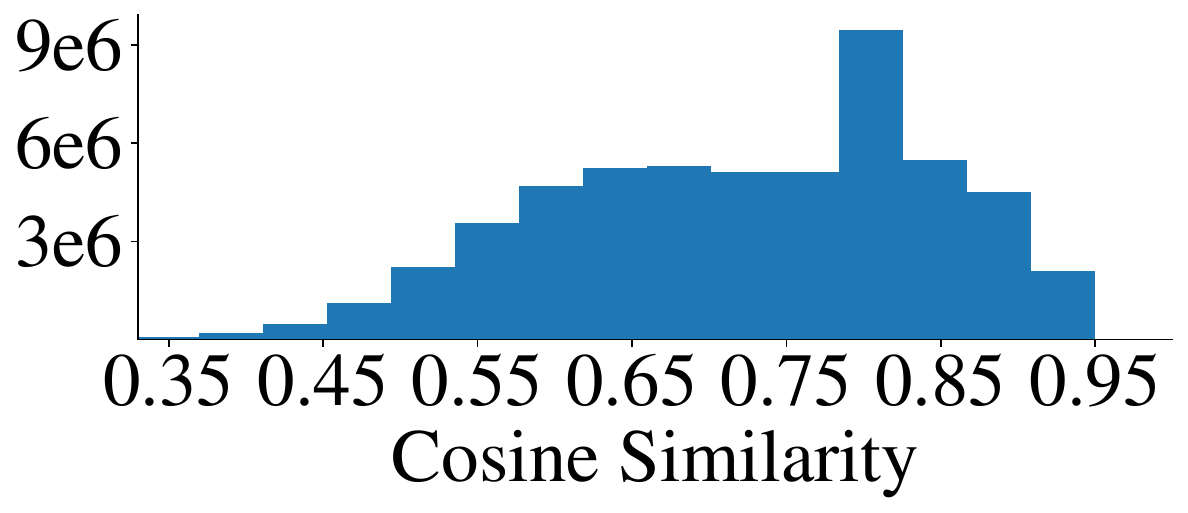}
  \end{minipage}
  \begin{minipage}{0.55\linewidth}
    \centering
    \includegraphics[width=1.02\linewidth]{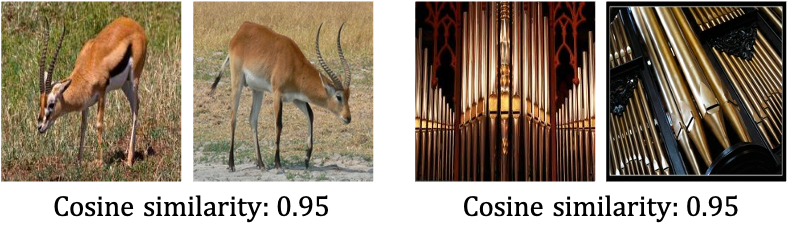}
  \end{minipage}
  \vspace{-0.4cm}
        \caption{\textbf{Filtering out pretraining data too similar to OVEN test/val.}}
         \label{fig:leak}
\end{figure}

\noindent \textbf{Preventing data leakage.}
Webli is already deduplicated against the train, val, and test splits of 68 common vision/vision-language datasets (see PaLI paper~\cite{pali2022}).
To be sure, in our paper, we further removed pretraining images with a cosine similarity (with CLIP-L/14 visual features) above 0.95 with any of the OVEN images.
We chose a 0.95 conservative threshold by looking at some examples: similarity 0.95 corresponds to conceptually similar images but clearly not duplicates (see Fig~\ref{fig:leak}).

\begin{algorithm}
\small
    \caption{\OURSf codes.}
    \label{alg}
    \KwData{Code length L, Text tokenizer $\Phi(.)$, Entities $\mathcal{E}$}
    
    \KwResult{$\mathcal{C} = \{c_e\}_{e \in \mathcal{E}}$}
    \For{$v \in [1, V]$}{
        $f_v = \frac{1}{\sum_{e \in \mathcal{E}} L_e}\sum_{e \in \mathcal{E}} \sum_{y^e_i \in \Phi(t_e)} \mathbb{1}_{y^e_i = v}$
    }
    $\mathcal{C} \leftarrow \varnothing $
    
    \For{$e \in \mathcal{E}$}{
     sort $\Phi(t_e)$ by decreasing frequencies:
     
     $\{y^e_{s_i}\}_{i \in [1, L_e]}$ such that $f_{y^e_{s_i}} \leq f_{y^e_{s_{i+1}}}$
     
         \For{$i \in [1, L - 1]$}{
            Set $c^e_i \leftarrow y^e_{s_i}$ 
         }
         $j = 0$
         
        \While{$c_e \in \mathcal{C}$ and $j \leq (L_e - L)$ }{
            $c^e_L \leftarrow y^e_{s_{L + j}}$
            
             $j = j + 1$
        }
        \While{$c_e \in \mathcal{C}$}{
            $c^e_L \leftarrow v' \sim [1, V]$
        }
        $\mathcal{C} \leftarrow \{c_e\} \bigcup \mathcal{C}$
    }
\end{algorithm}

\subsection{Finetuning on OVEN train set}
We finetune models on OVEN training set for 30,000 steps with a batch size of 256 and a learning rate of $1\text{e}^{-7}$.
Label smoothing is set at $0.1$.
Note that the finetuning schedule is relatively short (30,000 steps) because we observe that long finetuning (or equivalently, using a large learning rate) causes the model to forget about the unseen categories.

\subsection{Training on ImageNet-LT and Webvision}
\label{ap:sec:classif}
We train the model on ImageNet-LT and Webvision datasets with the batch size of 512 and a learning rate of $1\text{e}^{-4}$ for both encoder and decoder.
We do not use any label smoothing but apply a dropout of $0.1$.
We use $L=2$ for these experiments with \OURSf because unlike very large label-space this is enough not to resort to random tokens when ensuring that codes are unambious.

\begin{figure*}[t]
\centering
\includegraphics[width=0.9\linewidth]{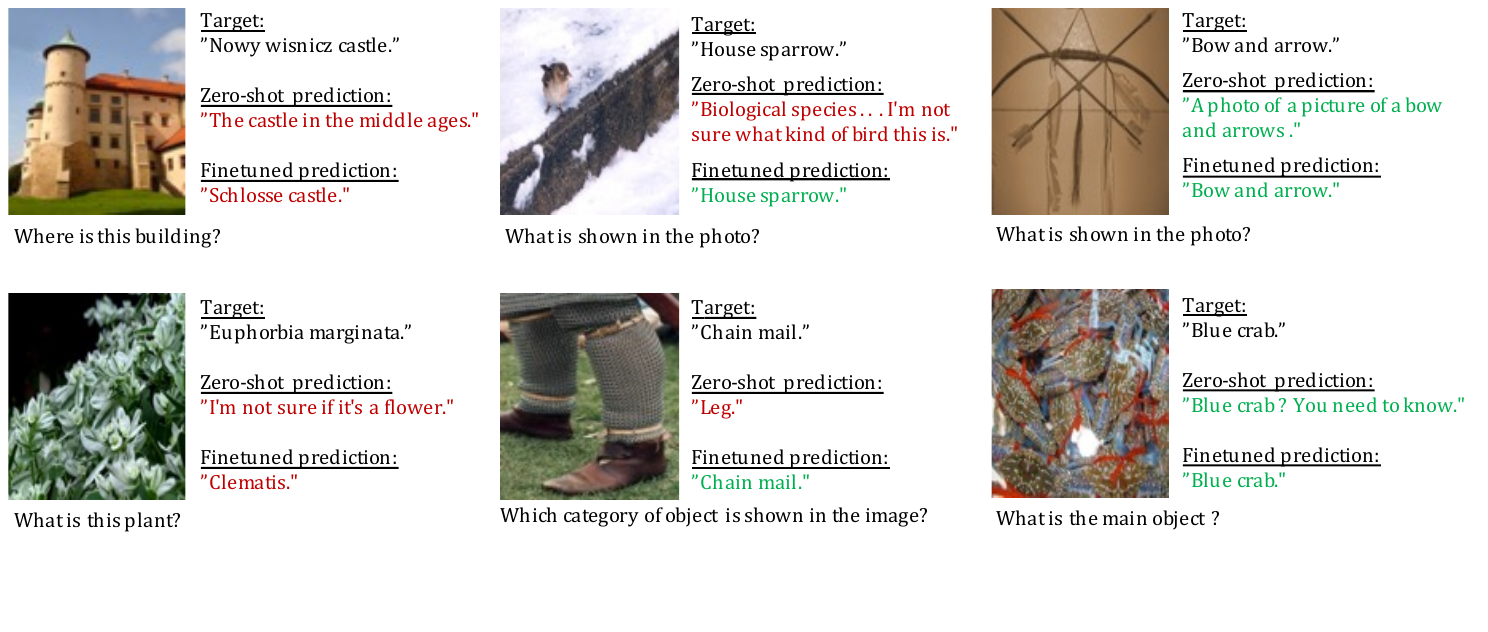}
\vspace{-1cm}
\caption{\textbf{Zero-shot versus finetuned captioning models predictions.}
We qualitatively compare the predictions of the captioning GiT-Large model when evaluated on OVEN in a zero-shot manner or after finetuning on OVEN train set.
}
\label{ap:fig:zeroshot_finetuned}
\end{figure*}

\subsection{Implementation details about the baselines}
\label{ap:sec:implem_baseline}
\paragraph{Dual encoder with CLIP-L/14.}
We use a learning rate of $3\text{e}^{-7}$, a batch size of $4096$ and we train for $200,000$ steps since training for longer deteriorates the performance.
During finetuning on OVEN training set, we find it important to still include some pretraining data: we randomly sample, with a probability of $90\%$, elements from the pretraining dataset.
Otherwise, when finetuning solely on OVEN, the model becomes too specialized for the seen categories and is not capable of discriminating between all the negative entities.
Alternatives could be to freeze some layers of the network during finetuning to prevent catastrophic forgetting.

\paragraph{\OURSa.}
We benchmark different values for the choice of $L$: \{$2$, $4$, $8$\} and $V$: \{$512$, $4096$, $32768$\} when using atomic codes.
Our default is to use $L=2$ and $V=4096$.
Note that this corresponds to more than $16$M possible different codes and we use only a subset of $6$M of those unique codes.

\paragraph{\OURSh.}
For HKC, we first represent each Wikipedia entity with a text embedding using the sentence-t5~\cite{ni2021sentence} encoder.
We have experimented with different ways of creating such embeddings, such as creating text embeddings from the Wikipedia titles, Wikipedia article summaries, and Wikipedia title and article summaries combined together.
We have observed that Wikipedia title and article summaries combined together produces the best embeddings.
We have tried different values of $k$: \{$10$, $100$, $1000$, $4096$, $8142$\}, and found that $k = 4096$ achieves the best performance in the validation set.

\section{More Experimental Results}
\subsection{Failure case analyses}
\begin{figure}[ht]
\small
 \begin{minipage}{0.4\linewidth}
    \centering
    \includegraphics[width=0.8\linewidth]{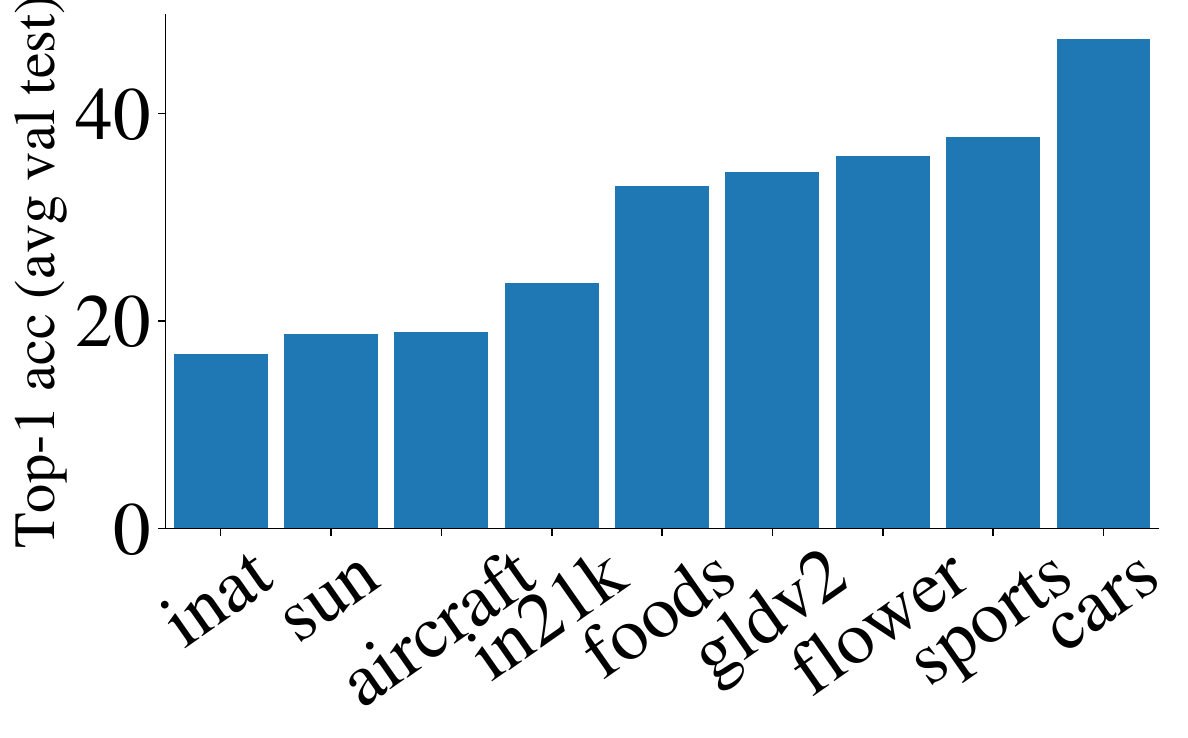}
  \end{minipage}\quad
  \begin{minipage}{0.55\linewidth}
    \centering
    \includegraphics[width=0.95\linewidth]{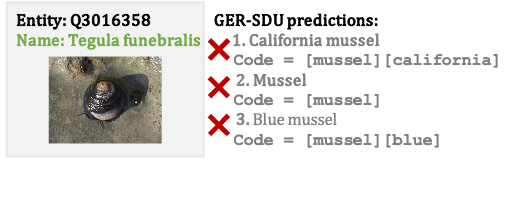}
  \end{minipage}
  \vspace{-0.4cm}
        \caption{\footnotesize{\textbf{(left): Accuracy per sub-task in OVEN. (right): A failure case example in iNaturalist (`inat') sub-task.}}}
         \label{fig:failure_cases}
\end{figure}

\begin{figure}[t]
\centering
\includegraphics[width=1.05\linewidth]{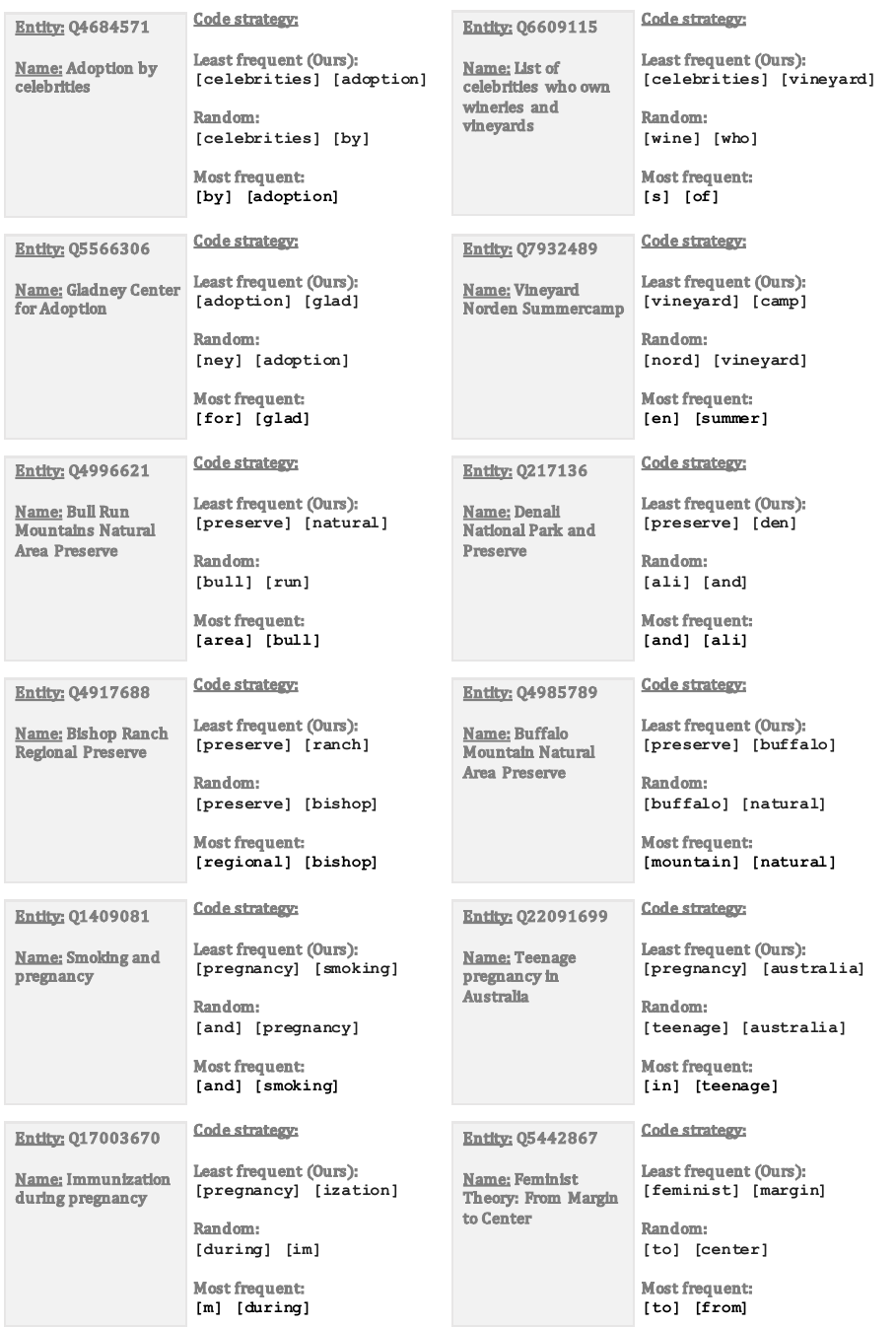}
\caption{\textbf{Token selection strategies in \OURSf.}
We qualitatively compare different alternative token selection strategies for \OURSf:
most frequent token or random token selection.
We use $L=2$ for this qualitative evaluation since this is easier to visually interpret that $L=4$, however the trends are consistent.
Quantitative evaluation is in the Table~\ref{fig:multi_ablat} of the main paper.
}
\label{ap:fig:ablation_sdu}
\end{figure}

We show in Fig.~\ref{fig:failure_cases} that our method works well across different entity types.
In many cases, our codes for animals (\eg \textit{`Glaucous winged gull'}), persons (\eg \textit{`List of celebrities who own wineries and vineyards'}) or organizations (\eg \textit{Gladney Center for Adoption'}) are interpretable, as shown in the qualitative examples in Fig~\ref{ap:fig:ablation_sdu} and Fig~\ref{ap:fig:vis}.
We see failure cases when semantics is difficult to infer from entity name alone.
This is often the case for scientific denomination of species as show in Fig.~\ref{fig:failure_cases}.
In future work, using external tools or Wikipedia page content could improve results in such cases.

\subsection{Zero-shot OVEN with captioning models}

\paragraph{Quantitative evaluation.}
In Table~\ref{ap:tab:captioning_zeroshot}, we transfer two captioning models, namely PALI-17B and GiT-Large~\cite{wang2022git}, both pre-trained on WebLI~\cite{pali2022} to the OVEN task.
We observe in Table~\ref{ap:tab:captioning_zeroshot} that these models transfer poorly in a zero-shot manner.
This can be explained by the major discrepancy between the pre-training captioning task (\ie describing an image with a caption) and the target entity recognition task.

\paragraph{Visual examples.}
We show some visual examples of predictions from the validation test between the zero-shot and finetuned GiT-Large models in Figure~\ref{ap:fig:zeroshot_finetuned} where we clearly see the difference of output between the zero-shot and finetuned GiT-Large models.
In the left column, we show two examples where both zero-shot and finetuned models fail.
However, even though the finetuned model fails to find the correct category of castle or plant, it still tries to output a fine-grained category of castle or plant.
This is not the case of the zero-shot model which gives a generic description of the entity, for example ``The castle in the middle ages.''.
In the middle column, we show examples where zero-shot fails but the finetuned model finds the correct category.
Finally, in the last column we show cases where the zero-shot model succeeds, but even when it does we observe that the generated caption is cluttered (for example with ``a photo of a picture'') while the finetuned model directly outputs the entity name.

Overall, the observation that models pre-trained from WebLi captions do not generalize well to OVEN entity recognition motivated us to create our entity-based pre-training described in Section~\ref{sec:training}.

\subsection{Entities with long names}
\begin{figure*}[t]
\centering
\includegraphics[width=1.05\linewidth]{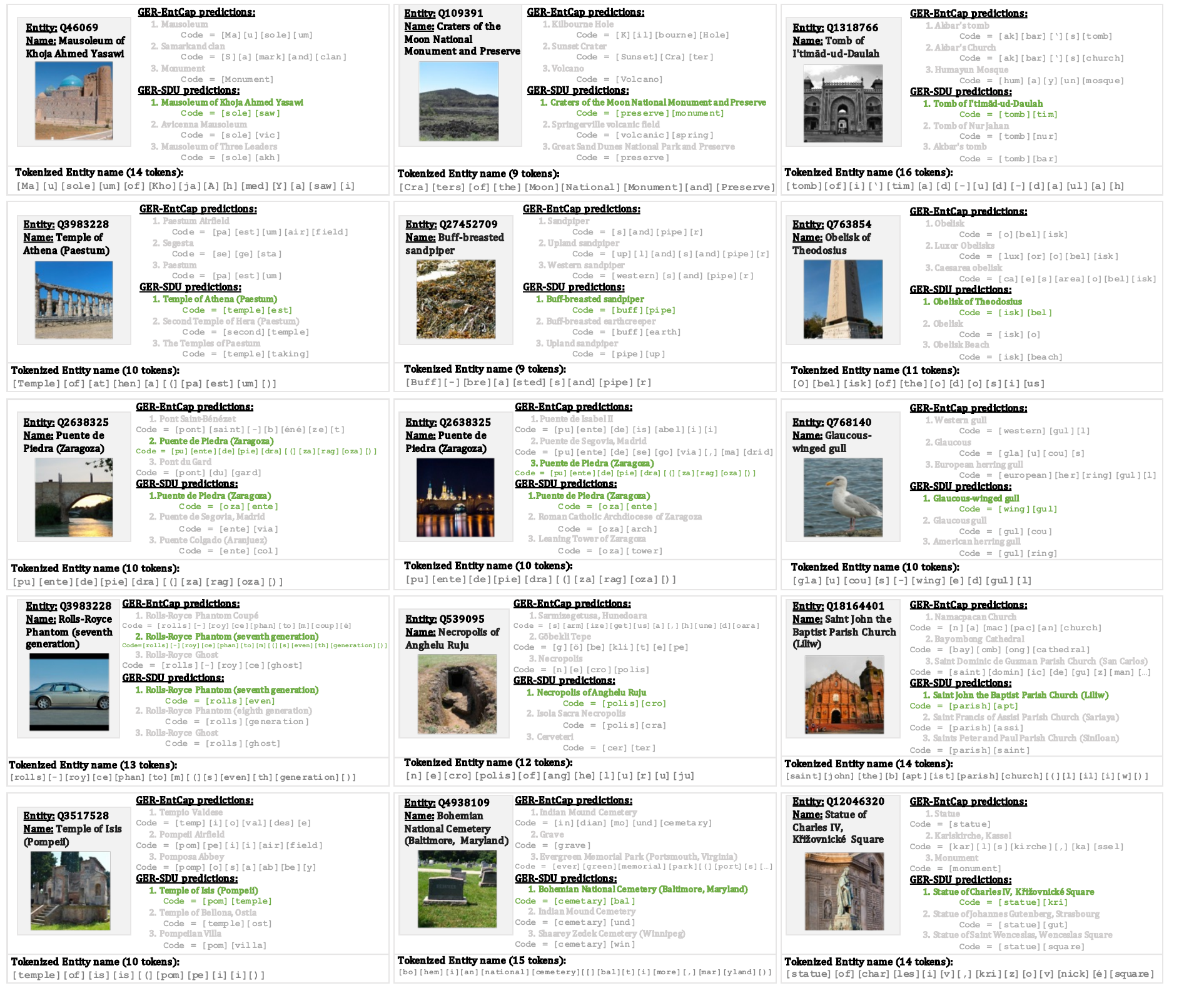}
\caption{\textbf{Qualitative study of \OURSf versus \OURSc.}
Visual examples of predictions for long entity name from 9 to 16 tokens.
For these visualizations with \OURSf, we use SentencePiece tokenizer~\cite{kudo2018sentencepiece} and $L=2$ in this evaluation since this leads to more visually interpretable codes.
Tokens are symbolized between brackets.
We report the top-3 predictions for \OURSf and for \OURSc codes, and color in green the correct predictions.
We observe that \OURSf codes are easier to predict as they contain less clutter than \OURSc codes.
Interestingly, we see that with \OURSf the top-3 predictions usually share a common token which re-group different semantically close entities.
}
\label{ap:fig:vis}
\end{figure*}

In Figure~\ref{ap:fig:vis}, we show more visual examples of \OURSf and \OURSc predictions for entities with long names.

\begin{figure*}[t]
\begin{minipage}{0.34\linewidth}
\centering
\small
\setlength{\tabcolsep}{1.5pt}
\begin{tabular}{@{}l ccccc@{}}
  \toprule
   Pretraining size (M) & 10.6 & 14.7 & 26.6 & 40.3 & 54.9 \\
  \midrule
  \OURSa & 0.9 & 6.8 & 11.4 & 13.8 & 15.3 \\
  \OURSf &  10.2 & 11.7 & 14.4 & 16.1 & 17.5 \\
 \rowcolor{Gray} Relative $\Delta$ (\%) & 1029 &  71 &  26 & 17 & 14 \\
  \bottomrule
\end{tabular}
\end{minipage}\quad
\begin{minipage}{0.22\linewidth}
\centering
\small
\setlength{\tabcolsep}{2pt}
\begin{tabular}{@{}l ccc@{}}
  \toprule
   Architecture size (M) & 114 & 179 & 397 \\
  \midrule
  \OURSa & 0.7 & 5.4 & 11.4 \\
  \OURSf & 5.6 & 9.5 & 14.4 \\
  \rowcolor{Gray} Relative $\Delta$ (\%) & 648 & 77 & 26 \\
  \bottomrule
 \end{tabular}
\end{minipage}\quad
\begin{minipage}{0.42\linewidth}
\centering
\small
\setlength{\tabcolsep}{3pt}
\begin{tabular}{@{}l ccccc@{}}
  \toprule
   \# entities (M) & 0.02 & 0.03 & 0.12 & 1.00 & 6.08 \\
  \midrule
  \OURSa & 34.6 & 34.0 & 29.7 & 21.5 & 11.4 \\
  \OURSf & 33.5 & 33.5 & 30.4 & 23.6 & 14.4 \\
  \rowcolor{Gray} Relative $\Delta$ (\%) & -3.1 & -1.6 & 2.2 & 9.6 & 26.4 \\
  \bottomrule
\end{tabular}
\end{minipage}\quad
\captionof{table}{\textbf{Semantically-structured (\OURSf) versus unstructured (\OURSa) codes.}
We report the numbers corresponding to Figure~\ref{fig:size_label_space} of the main paper.
The pretraining dataset sizes of 10.6M, 14.7M, 26.6M, 40.3M and 54.9M correspond respectively to setting $k$ to 2, 5, 20, 50 and 100.
The architecture sizes with 114M, 179M and 397M parameters correspond respectively to variant Small, Base and Large of the model.
The label space sizes with 20549, 30549, 120549, 1000549 and 6084491 different entities correspond respectively to having 0, 10k, 100k, 1M and 6M entities acting as distractors.
}
\label{ap:tab:size_label_space}
\end{figure*}

\subsection{Different token selection strategies}
In Figure~\ref{ap:fig:ablation_sdu}, we show visual examples of codes generated with alternatives of selecting the least frequent token in \OURSf.
We compare with selecting instead the most frequent token and with selecting a random token in the entity name.
Quantitative evaluation is in Table~\ref{fig:multi_ablat} of the main paper.
In Figure~\ref{ap:fig:ablation_sdu}, we observe that codes generated with least frequent token strategy are the most semantically structured.
Indeed, in this case the entities ``Adoption by celebrities'' and ``List of celebrities who own wineries and vineyards.'' share a common token (the token corresponding to ``celebrities'') while there is no intersection of token between those two entities for the most frequent or the random strategies.
We observe the same effect across several group of entities that we intuitively expect to have shared tokens, for example with ``smoking and pregnancy'', ``teenage pregnancy in Australia'' and ''Immunization during pregrancy'', or with ``Denall National Park and Preserve'' and ``Bishop Ranch Regional Preserve''.

\subsection{Numbers corresponding to Fig.~\ref{fig:size_label_space} main paper}
We report in Table~\ref{ap:tab:size_label_space} the numbers corresponding to the experiments shown in Figure~\ref{fig:size_label_space} of the main paper.

\end{document}